\documentclass[10pt,twocolumn,letterpaper]{article}

\usepackage[T1]{fontenc}

\usepackage[pagenumbers]{wacv}   %

\usepackage{cuted}
\usepackage{capt-of}

\usepackage{newtxtext}
\usepackage{amsmath,amssymb}
\usepackage{multirow}
\usepackage{makecell}
\usepackage{siunitx}
\usepackage{pgfplots}
\usepackage{bbm}
\usepackage{algorithm}
\usepackage{algpseudocode}
\usepackage{multicol}   %
\usepackage{wrapfig}    %

\pgfplotsset{compat=1.18}
\newcommand{\method}{\textsc{AttnSteer}\xspace}

\newcommand{\vx}{\mathbf{x}}
\newcommand{\vz}{\mathbf{z}}

\newcommand{\attn}{A}                      %

\newcommand{\blockbest}[1]{\textbf{\boldmath#1}}
\newcommand{\allbest}[1]{\underline{#1}}
\newcommand{\up}{$\uparrow$}
\newcommand{\down}{$\downarrow$}

\definecolor{linkblue}{rgb}{0.21,0.49,0.74}
\usepackage[pagebackref,breaklinks,colorlinks,allcolors=linkblue]{hyperref}
\usepackage{bookmark}

\def\wacvPaperID{2168}
\def\confName{WACV}
\def\confYear{2027}
\title{Will It Render? Learned Attention Steering for Object Presence in Text-to-Video}

\author{%
Ben Fishman\textsuperscript{1,$*$} \quad\quad
Dvir Samuel\textsuperscript{1,2} \quad\quad
Gal Chechik\textsuperscript{1,2} \quad\quad
Idan Schwartz\textsuperscript{1}\\[4pt]
\textsuperscript{1}Bar-Ilan University \quad\quad \textsuperscript{2}NVIDIA
}

\definecolor{wacvblue}{rgb}{0.21,0.49,0.74}
\definecolor{backcolour}{rgb}{0.95,0.95,0.92}
\colorlet{reddish}{red!70}

\begin{document}
\setcounter{tocdepth}{4}
\maketitle
{\renewcommand\thefootnote{}\footnotetext{Accepted to WACV 2027.
\textcopyright~2027 IEEE\@. Personal use of this material is permitted.
Permission from IEEE must be obtained for all other uses, in any current or
future media, including reprinting/republishing this material for advertising or
promotional purposes, creating new collective works, for resale or
redistribution to servers or lists, or reuse of any copyrighted component of
this work in other works. $^*$ \texttt{benf22@gmail.com}}}

\setlength{\stripsep}{2pt}
\begin{strip}
  \centering
  \captionsetup{hypcap=false}
  \includegraphics[width=\textwidth]{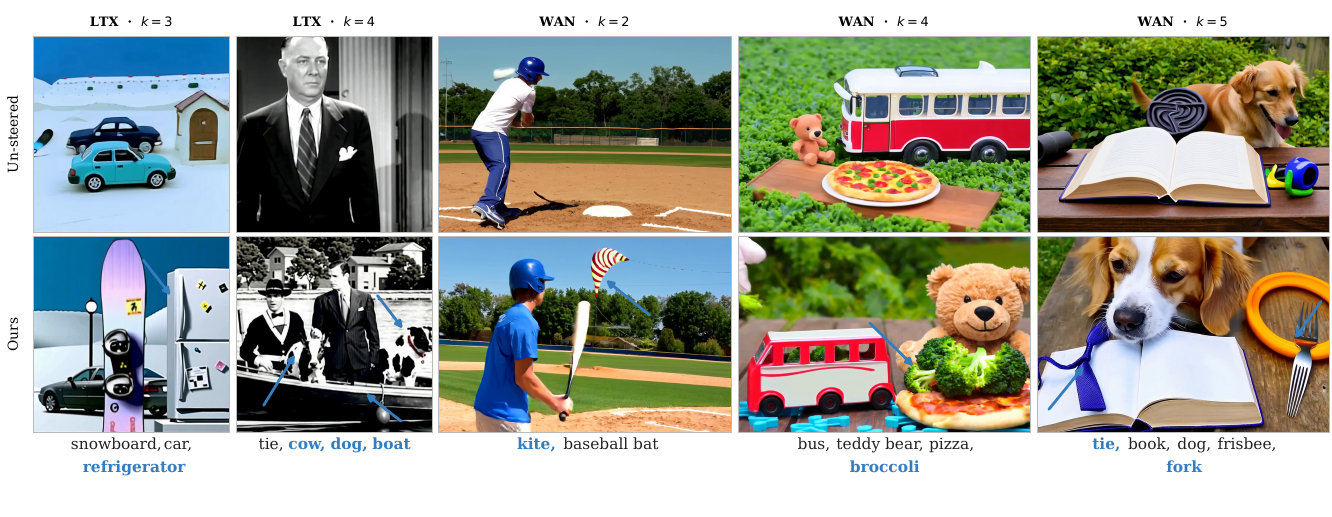}
    \captionof{figure}{\textbf{Steering object presence in text-to-video generation.}
    Mid-clip snapshots from LTX-2 and Wan~2.1 for prompts containing $k$ requested
    objects; both rows share the prompt and the seed. The un-steered backbones (top) frequently neglect requested objects,
    highlighted in blue text below, whereas inference-time steering with \method{} (bottom)
    recovers these missing objects while preserving video quality. Blue arrows show the recovered objects. Object neglect
    becomes increasingly severe as the number of requested objects grows.
    }
  \label{fig:teaser}
\end{strip}

\begin{abstract}
Text-to-video diffusion models generate increasingly realistic videos, yet often fail to render all objects requested by the prompt, particularly as the number of objects grows. Improving such fine-grained adherence through training requires substantial video supervision, motivating inference-time methods that steer intermediate diffusion latents. Existing approaches typically define hand-crafted objectives over cross-attention maps, relying on indirect attention statistics rather than whether the model ultimately renders a requested object. We propose \method{}, a learned attention-steering model that predicts object rendering directly from intermediate cross-attention. \method{} is trained entirely on the generator's own trajectories using automatically obtained detector labels and contains only 37k parameters. At inference time, its predictions define differentiable objectives for steering object presence; the same framework also extends to frame-level temporal control. \method{} improves all-object accuracy over the strongest baseline from $51.1\%$ to $75.5\%$ on LTX-2, from $40.4\%$ to $59.9\%$ on Wan~2.1, and from $56.1\%$ to $65.9\%$ on CogVideoX, and raises temporal specificity from $35.7\%$ to $52.6\%$, at comparable video quality.
Project page: \href{https://benf22.github.io/will-it-render/}{https://benf22.github.io/will-it-render/}
\end{abstract}

\addtocontents{toc}{\protect\setcounter{tocdepth}{-1}}%
\section{Introduction}
\label{sec:intro}

Text-to-video diffusion transformers~\cite{peebles2023dit} produce increasingly convincing videos, yet remain unreliable at fine-grained prompt adherence. In particular, as prompts request more objects, models increasingly fail to render all of them. This failure is especially limiting for applications such as simulation and synthetic-data generation, where missing even one requested entity can make a generated clip unusable. We therefore focus on \emph{object presence}: ensuring that each object requested by the prompt is actually rendered. We additionally study temporal control as a secondary setting, asking whether the same steering principle can determine \emph{when} an object appears.

One way to improve adherence is to train on additional constrained examples. For video, however, obtaining such supervision is expensive, while the space of possible object combinations is difficult to cover exhaustively. Inference-time steering offers an alternative: rather than retraining the generator, it modifies the noisy latent during sampling using signals already available within the model. Cross-attention is particularly useful because it captures rich interactions between prompt tokens and the evolving spatio-temporal representation~\cite{schwartz2017high,schwartz2019factor}. Existing methods typically reduce these features to hand-crafted statistics, such as maximizing the peak attention of a neglected object~\cite{chefer2023attendexcite} or matching attention to a prescribed temporal pattern~\cite{tempocontrol}.

Such statistics are only proxies for the quantity we ultimately care about: whether the requested object will actually be rendered. Moreover, reducing high-dimensional cross-attention to a predefined statistic can discard predictive structure distributed across layers, heads, space, and time. In image generation, Learn-to-Steer~\cite{learn2steer} showed that hand-crafted attention objectives can be replaced by a learned classifier for spatial composition. Similarly, HEaD~\cite{betti2024head,betti2026head} predicts from cross-attention whether the requested objects will appear in an image, and restarts the generation when they will not. Extending learned steering to video, however, poses two practical challenges: obtaining suitable video supervision and keeping the steering model inexpensive enough to evaluate repeatedly during an already costly generation process.

We address these challenges with \method{}, a learned attention-steering model that predicts the eventual rendering outcome directly from intermediate cross-attention. We generate training clips using the frozen video model itself and automatically label whether each requested object is rendered, requiring no external video supervision. \method{} operates on the generator's per-layer, per-head attention features and contains only 37k parameters. At inference time, its prediction defines a differentiable object-presence objective whose gradient is propagated back to the video latent. The same formulation applies frame by frame, allowing the learned objective to support temporal control as well.

Multi-object prompts introduce an additional difficulty: independently steering several requested objects can produce conflicting latent updates. We therefore introduce a conflict-aware gradient aggregation rule that normalizes each object's gradient, anchors the update on the object least likely to be rendered, and suppresses locally opposing gradient contributions.

Across LTX-2, Wan~2.1 and CogVideoX, \method{} improves object presence. Under a three-detector majority vote, all-object accuracy improves by $37.2$, $32.8$, and $35.4$ points over the un-steered LTX-2, Wan~2.1, and CogVideoX backbones, and by $37.7$, $20.7$, and $9.8$ points over Attend-and-Excite~\cite{chefer2023attendexcite}, with larger gains as the number of requested objects increases. 
We further find that the same learned objective transfers to temporal steering, improving $F1$ score by $2$ points and outperforming TempoControl~\cite{tempocontrol} on interval-based metrics. These gains come at comparable VBench video quality.

\paragraph{Contributions.}
\begin{itemize}\itemsep0pt
    \item We introduce \method{}, an outcome-based inference-time steering objective that learns from cross-attention whether a requested object will ultimately be rendered.

    \item We develop a self-generated supervision scheme that trains the steering model entirely from the generator's own trajectories and automatic outcome labels, without external video supervision.

    \item We introduce conflict-aware multi-object steering that normalizes per-object gradients and locally suppresses opposing updates.
\end{itemize}

\section{Related Work}
\label{sec:related}

\paragraph{Video generation and fine-grained adherence.}
Diffusion models~\cite{ho2020denoising,rombach2022highresolution} established a powerful framework for conditional image generation and were subsequently extended to video~\cite{singer2023makeavideo,blattmann2023videoldm}. As video quality improved, research increasingly addressed finer aspects of conditioning, including alignment with audio~\cite{yariv2024diverse,rahamim2027reccar}, object trajectories~\cite{wang2024boximator,geng2025motionprompting}, spatio-temporal layouts~\cite{lian2024lvd,tian2024videotetris}, and event timing~\cite{wu2025mind,xu2026switchcraft}. Yet video models can still fail at a more basic aspect of prompt adherence: rendering all requested objects, particularly as the number of objects increases~\cite{sun2025t2vcompbench}. Our work focuses directly on this object-presence failure.

\paragraph{Inference-time attention optimization.}
Inference-time optimization improves generation without retraining the base model by updating the diffusion trajectory using classifier gradients~\cite{dhariwal2021diffusion}, off-the-shelf guidance functions~\cite{bansal2023universal}, conditioning embeddings~\cite{schwartz2023discriminative,zafar2026detection}, or newly introduced test-time parameters~\cite{qu2026ttom,shpitzer2026siso}. Cross-attention is particularly useful because it exposes spatial correspondences between prompt concepts and generated content~\cite{hertz2022prompt}, and has been directly optimized to improve semantic and spatial adherence~\cite{chefer2023attendexcite,li2023divide,rassin2023syngen,meral2024conform,marioriyad2025}. Likewise, attention relevance maps have been optimized to focus classifiers on the foreground object~\cite{chefer2022optimizing}. This idea extends naturally to video: LVD~\cite{lian2024lvd} optimizes attention against spatio-temporal layouts, Peekaboo~\cite{jain2024peekaboo} masks attention to follow them, Vico~\cite{yang2024vico} equalizes the influence of each prompt token on the video, and TempoControl~\cite{tempocontrol} uses attention objectives for temporal constraints. Most such approaches rely on hand-crafted objectives over attention. Readout Guidance~\cite{luo2024readout} instead trains lightweight heads on diffusion features and steers sampling toward their targets, Learn-to-Steer~\cite{learn2steer} learns an attention-based objective for spatial relations in images, and CVG~\cite{shaulov2026cvg} learns compositional-relation guidance for video. Closest to our predictor, HEaD~\cite{betti2024head,betti2026head} predicts from cross-attention whether all requested objects will appear in an image and restarts the generation with a new seed otherwise, and Diffusion Probe~\cite{diffusionprobe2026} predicts the quality of the final image from early cross-attention. In contrast, \method{} turns the predicted rendering outcome of each requested object into a differentiable steering objective for video, learned from self-generated trajectories and automatic outcome labels.

\section{Method}
\label{sec:method}

Our method has two stages.
First, we train a model to predict whether a requested object will ultimately
be rendered from intermediate cross-attention features
(\cref{fig:training}).
Second, during generation, we use this prediction as a differentiable
inference-time objective and optimize the video latent while keeping the
generator fixed (\cref{fig:method}).

\subsection{Preliminaries}
\label{sec:method:prelim}

\paragraph{Text-to-video diffusion.}
A text-conditioned video diffusion model $D_t$ iteratively denoises a latent
\begin{equation}
    \vz_t
    \in
    \mathbb{R}^{C\times f\times h\times w},
    \label{eq:latent}
\end{equation}
where $t\in\{0,\ldots,T\}$ is the denoising step, $C$ is the latent channel
dimension, and $f\times h\times w$ is the temporal and spatial latent grid. In our notation, $t=0$ is the noisy signal and $t=T$ is the clean video.

Let the prompt contain $m$ text tokens.
For transformer layer $\ell\in\{1,\ldots,L\}$ and attention head
$r\in\{1,\ldots,H\}$, let
\begin{equation}
    A_{\ell,r}(\vz_t)
    \in
    \mathbb{R}^{(fhw)\times m}
\end{equation}
denote the cross-attention matrix from latent positions to prompt tokens.

For a target object $O_i$, let
$\mathcal{T}_i\subseteq\{1,\ldots,m\}$ denote its corresponding prompt-token
indices.
We sum the attention assigned to these tokens,
\begin{equation}
    A_{i,\ell,r}(\vz_t)[q]
    =
    \sum_{j\in\mathcal{T}_i}
    A_{\ell,r}(\vz_t)[q,j],
    \label{eq:object_attention}
\end{equation}
where $q\in\{1,\ldots,fhw\}$ indexes a spatio-temporal latent position.
Stacking these maps across layers and heads gives the object-specific attention
tensor
\begin{equation}
    \attn_i(\vz_t)
    \in
    \mathbb{R}^{L\times H\times f\times h\times w}.
    \label{eq:attn_features}
\end{equation}

In addition to $\attn_i(\vz_t)$, \method{} receives the mean text key associated
with $\mathcal{T}_i$ at each layer and head.
We suppress this auxiliary conditioning from the notation for clarity.

\paragraph{Inference-time attention steering.}
Let $\mathcal{S}\subseteq\{0,\ldots,T\}$ denote the denoising steps at which
steering is applied. At a step $t\in\mathcal{S}$, starting from the
sampler latent $\vz_t^{(0)}$, steering performs $N$ gradient
updates on an objective $\mathcal{L}$,
\begin{equation}
    \vz_t^{(n+1)}
    =
    \vz_t^{(n)}
    -
    \eta
    \nabla_{\vz_t^{(n)}}
    \mathcal{L}\!\left(\vz_t^{(n)}\right),
    \qquad n=0,\ldots,N-1,
    \label{eq:generic_steering}
\end{equation}
where $\eta$ is the step size. The optimized latent is then returned to the
sampler and denoising continues.

Existing approaches typically construct $\mathcal{L}$ from hand-crafted
statistics of cross-attention. For example,
Attend-and-Excite~\cite{chefer2023attendexcite} encourages an object by
minimizing
\begin{equation}
    \mathcal{L}_{\mathrm{A\&E}}
    =
    1-\max_q \bar A_i(q),
    \label{eq:attend_excite}
\end{equation}
where $\bar A_i$ is its aggregated attention map for object $O_i$.
Rather than prescribing which attention statistic should indicate object
presence, we learn an objective that predicts whether the requested object will be rendered.
\subsection{Learning an Object-Presence Objective}
\label{sec:method:attnguide}

\begin{figure}[t]
  \centering
  \includegraphics[width=\linewidth]{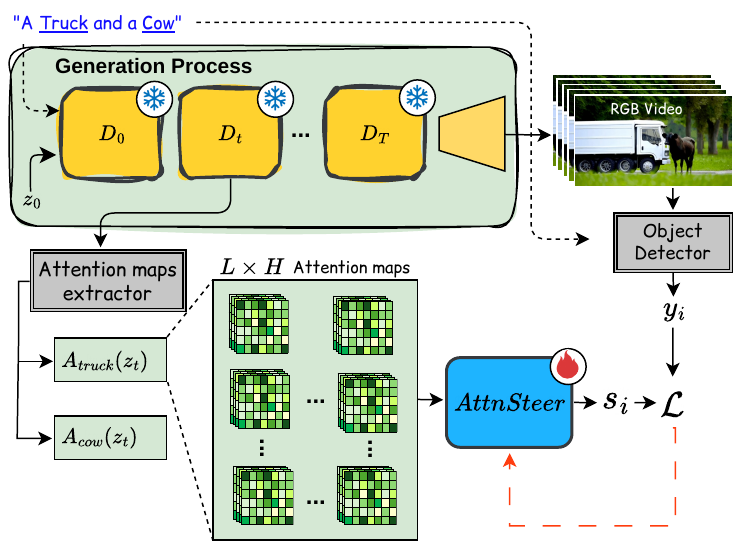}
  \caption{\textbf{Training \method{} with self-generated supervision.}
  During sampling, object-specific cross-attention features are extracted from
  the frozen generator and paired with detector-derived labels from the final
  generated video. The rendering loss trains \method{} to answer the
  video-level question: \textit{will it render?}}
  \label{fig:training}
\end{figure}

Our key idea is to replace a hand-crafted attention objective with a learned
predictor of the outcome we care about: \emph{will the requested object be rendered?}
We train an independent predictor $\method{}^{(t)}$ for each captured step $t$, and a steered step without one uses the nearest predictor. For object $O_i$, the predictor outputs a rendering score
\begin{equation}
    s_i(\vz_t)
    =
    \method{}^{(t)}
    \!\left(
        \attn_i(\vz_t)
    \right),
    \label{eq:presence_prediction}
\end{equation}
where $s_i(\vz_t)\in\mathbb{R}$ is a logit predicting whether $O_i$ will be
rendered somewhere in the final video.

For a target $y\in\{0,1\}$, the rendering loss is
\begin{equation}
    \mathcal{L}^{\mathrm{render}}_i(\vz_t;y)
    =
    \operatorname{BCE}
    \!\left(
        s_i(\vz_t),y
    \right),
    \label{eq:render_loss}
\end{equation}
where $\operatorname{BCE}$ denotes binary cross-entropy with logits.
At inference time, $y=y_i^\star$ is the desired rendering outcome. For a
requested object, we set $y_i^\star=1$ and backpropagate
$\mathcal{L}^{\mathrm{render}}_i(\vz_t;y_i^\star)$ through the frozen predictor
and generator to $\vz_t$. Thus, rather than prescribing what a successful
attention map should look like, we steer toward attention patterns that predict
the requested object will ultimately be rendered.

\paragraph{Self-generated supervision.}
The remaining question is how to train the predictor without an external
labeled video dataset. As illustrated in \cref{fig:training}, we generate
videos using the frozen generator and record $\attn_i(\vz_t)$ for every target
object at each $t\in\mathcal{S}$ along the same sampling trajectory that
produces the final video.

For each object $O_i$ we apply three object detectors to the decoded frames and
take their majority vote per frame, discarding frames without a reliable
majority. The clip-level label $y_i$ marks the object as rendered when it is
detected in at least $15\%$ of the scored frames, so that a single stray
detection does not make an object present; \cref{sec:app:evaluation} gives the
frame counts and how sensitive the labels are to them.
We train the predictor using the same rendering loss,
$\mathcal{L}^{\mathrm{render}}_i(\vz_t;y_i)$.

Each intermediate attention representation is therefore paired with the
eventual rendering outcome of the same generation. No external video
supervision or inversion of real videos is required.

We construct prompts containing between one and seven objects, and split the data by prompt identity
so that the same prompt never appears in both training and validation.
Full prompt construction, detector labeling, filtering, and balancing details
are provided in \cref{sec:app:method}.

\paragraph{\method{} architecture.}
\method{} maps the object-specific attention tensor
$\attn_i(\vz_t)\in\mathbb{R}^{L\times H\times f\times h\times w}$ to the
rendering score $s_i(\vz_t)$. We keep the $H$ attention heads separate and
learn how to combine information across the $L$ layers. A grouped
$1{\times}1{\times}1$ convolution produces $K=2$ learned layer combinations
for each head,
\begin{equation}
    \widetilde A_i
    =
    \operatorname{LayerAgg}
    \!\left(\attn_i(\vz_t)\right)
    \in
    \mathbb{R}^{2H\times f\times h\times w}.
    \label{eq:layer_aggregation}
\end{equation}

We then process each frame independently using a small ResNet
variant~\cite{he2016deep}, consisting of a convolutional stem followed by three
spatial residual blocks. A final temporal residual block mixes information
across frames. Global average pooling over space and time followed by a linear
projection produces the rendering score,
\begin{equation}
\begin{aligned}
    F_i
    &=
    R_{\mathrm{temp}}
    \!\left(
        \operatorname{ResNet}(\widetilde A_i)
    \right), \\
    s_i(\vz_t)
    &=
    \operatorname{Linear}
    \!\left(
        \operatorname{AvgPool}(F_i)
    \right).
\end{aligned}
\label{eq:predictor_architecture}
\end{equation}

This architecture is designed for efficiency. A small architecture ablation found this capacity sufficient for
predicting rendering outcomes. Full architectural and training details are
provided in \cref{subsec:app:impl}.

\begin{figure}[t]
  \centering
  \includegraphics[width=1.0\linewidth]{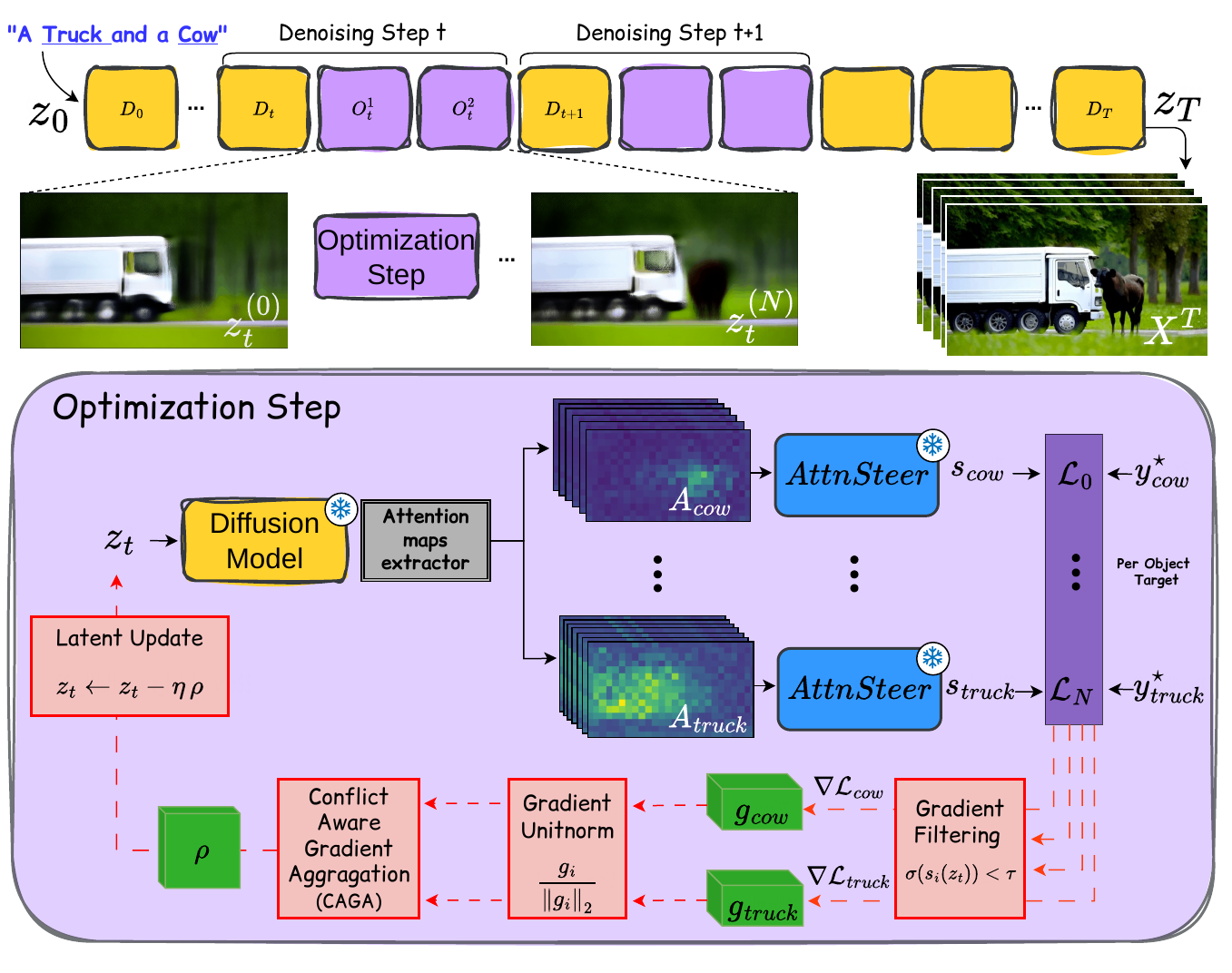}
  \caption{\textbf{Inference-time steering with \method{} for the prompt
  ``A Truck and a Cow.''}
  At each selected denoising step, object-specific cross-attention features are
  scored by the corresponding frozen \method{} predictor. The resulting
  per-object losses are differentiated with respect to the video latent.
  For multi-object prompts, their normalized gradients are combined using our
  conflict-aware gradient aggregation (CAGA) which outputs $\rho$ that is used to update the latent. This
  optimization is repeated for $N$ iterations before denoising resumes. }
  \label{fig:method}
\end{figure}

\paragraph{Steering with the learned objective.}
\Cref{fig:method} summarizes the inference procedure. We use the
inference-time optimization described in \cref{sec:method:prelim}, replacing
the hand-crafted attention objective with our learned rendering loss. At
optimization iteration $n$, object $O_i$ therefore induces the gradient
\begin{equation}
    g_i^{(n)}
    =
    \nabla_{\vz_t^{(n)}}
    \mathcal{L}^{\mathrm{render}}_i
    \!\left(
        \vz_t^{(n)};y_i^\star
    \right),
    \label{eq:object_gradient}
\end{equation}
with $y_i^\star=1$ for a requested object. Gradients propagate through
\method{} and the frozen generator to the latent. For a single object, this
gradient can be used directly in the steering update; for multiple objects, we
combine the resulting object-specific gradients as described next.
\subsection{Multi-Object Steering}
\label{sec:method:multi}

When a prompt requests $M$ objects, each object produces its own steering
gradient $g_i^{(n)}$ at optimization iteration $n$. Their magnitudes are not
directly comparable, so before combining them we normalize each gradient,
\begin{align}
    \hat g_i^{(n)}
    =
    \frac{g_i^{(n)}}{\left\lVert g_i^{(n)} \right\rVert_2+\epsilon},
    \qquad i=1,\ldots,M,    \label{eq:grad_norm}
\end{align}
where $\epsilon$ is a small numerical constant. Thus, no object
dominates the update because of its gradient norm.

The remaining problem is directional conflict: averaging two normalized
gradients can still cancel useful updates, a problem also studied in multi-task learning~\cite{yu2020pcgrad,liu2021cagrad} and addressed gradient-free in FK-ReMask~\cite{mousai2026fkremask} via particle selection. Our aggregation therefore gives
priority to the object that \method{} currently considers least likely to be
rendered, i.e., the object with the lowest rendering logit
$s_i(\vz_t^{(n)})$, whose gradient initializes the steering direction.

We then consider the remaining objects one at a time. At a spatio-temporal
latent position $q$, both the current steering direction and an object's
gradient are vectors in $\mathbb{R}^{C}$ over the latent channels. We add the
new gradient at $q$ only if its inner product with the direction accumulated
there is non-negative. In other words, an object may reinforce the current
update, but is not allowed to locally reverse it. This procedure is summarized
in \cref{alg:gradient_aggregation}.
The procedure applies only to objects that \method{} does not yet predict to be rendered, i.e., those with $\sigma(s_i(\vz_t^{(n)}))<\tau$.
When all objects exceed the threshold, the update is either skipped entirely
(\emph{zero} fallback) or applied to the lowest-scoring object (\emph{worst}
fallback), depending on the configuration, as listed in \cref{tab:hparams}.

\begin{algorithm}[t]
\caption{Conflict-aware gradient aggregation}
\label{alg:gradient_aggregation}
\begin{algorithmic}[1]
\Require Normalized object gradients $\hat g_i^{(n)}$ and rendering scores
$s_i(\vz_t^{(n)})$
\State Keep only objects with $\sigma(s_i(\vz_t^{(n)}))<\tau$
\State Order objects from lowest to highest rendering score
\State $d \gets$ gradient of the lowest-scoring object
\For{each remaining object $i$}
    \For{each latent position $q$}
        \If{$\langle \hat g_i^{(n)}[q], d[q]\rangle \geq 0$}
            \State $d[q] \gets d[q] + \hat g_i^{(n)}[q]$
        \EndIf
    \EndFor
\EndFor
\State \Return $\rho^{(n)} \gets d/M$
\end{algorithmic}
\end{algorithm}

\begin{figure}[t]
  \centering
  \includegraphics[width=\linewidth]{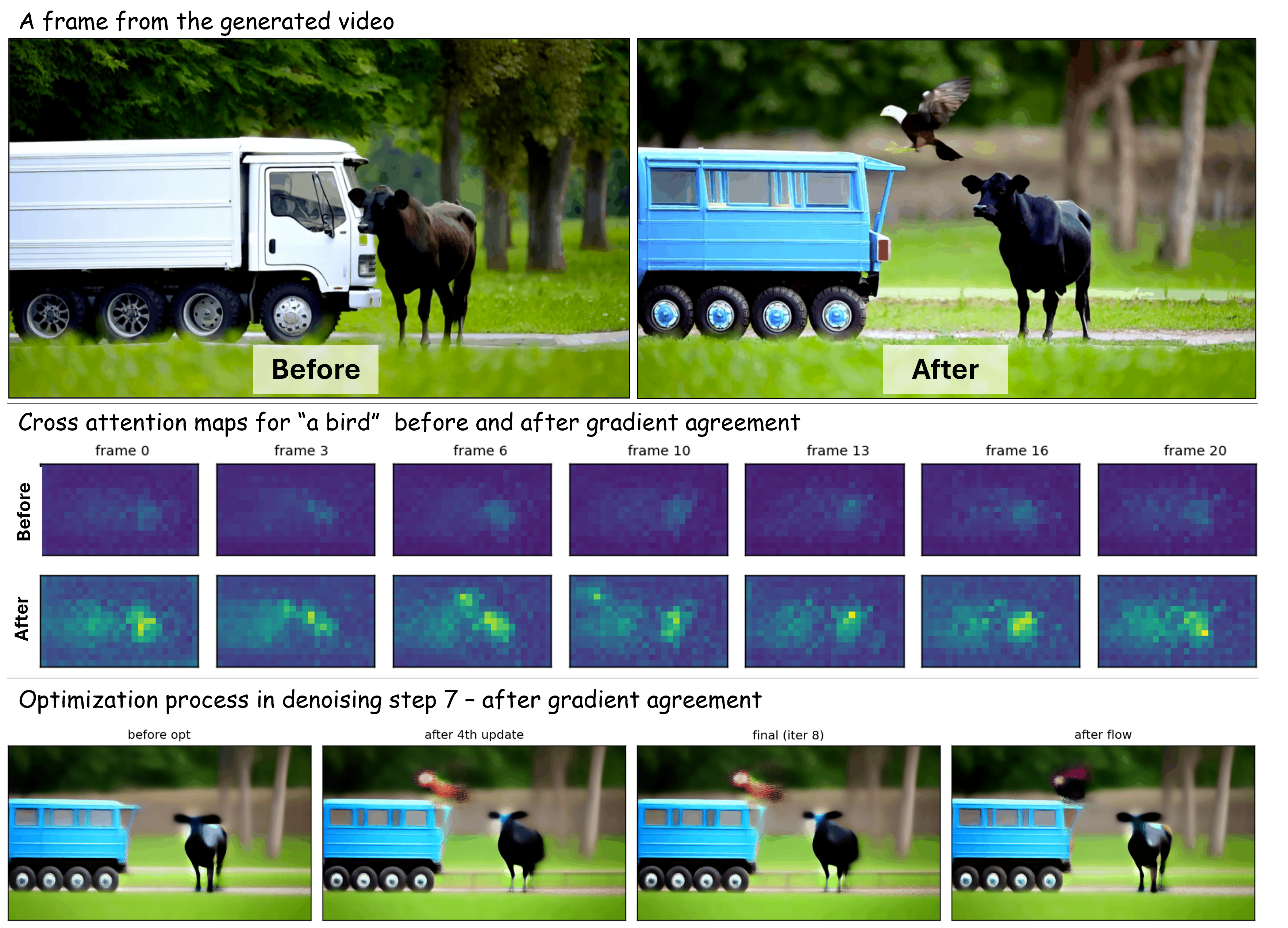}
  \caption{\textbf{Effect of conflict-aware gradient aggregation.}
  (a) Plain gradient averaging fails to render the requested bird, whereas our
  aggregation introduces it.
  (b) The corresponding cross-attention maps for ``a bird.''
  (c) Predicted denoised samples show the bird emerging during steering.}
  \label{fig:gradient_agreement}
\end{figure}

The division by $M$ keeps the update scale comparable as the number of
requested objects changes. \Cref{fig:gradient_agreement} illustrates the
effect: plain averaging fails to introduce the missing bird, whereas
\cref{alg:gradient_aggregation} suppresses conflicting contributions and the
bird emerges during optimization.

Finally, we use $\rho^{(n)}$ in place of the single-object gradient in the
latent update,
\begin{equation}
    \vz_t^{(n+1)}
    =
    \vz_t^{(n)}
    -
    \eta\,\rho^{(n)}.
    \label{eq:full_steering}
\end{equation}
For a single requested object, the procedure reduces to its normalized
gradient.

\begin{figure*}[t]
    \centering
    \includegraphics[width=\linewidth]{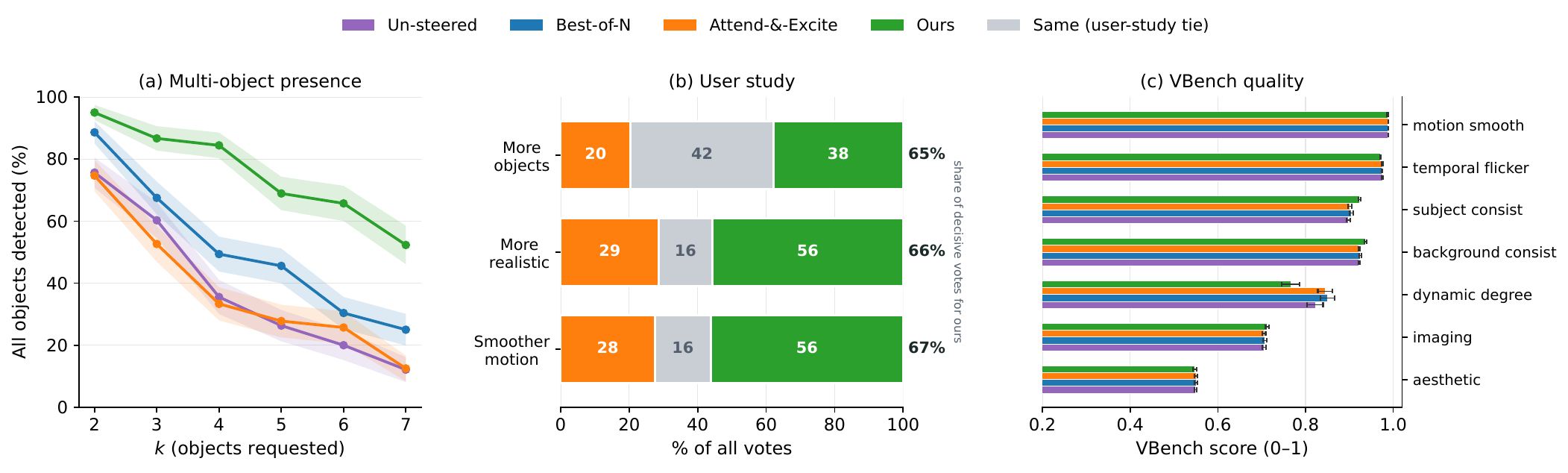}
    \vspace{-3mm}
    \caption{\textbf{LTX-2 results.}
    \textbf{(a) Multi-object presence:} all-object accuracy as the number of requested
    objects $k$ increases; our method is favored throughout.
    \textbf{(b) User study:} human annotators also prefer our method over the
    Attend-and-Excite baseline.
    \textbf{(c) Quantitative quality:} VBench quality per dimension shows that steering
    causes almost no quality trade-off --- ours is on par with the baselines on every
    dimension except dynamic degree.
    The four methods keep the same color across panels; bands and error bars denote
    $\pm1$ standard error. Wan~2.1 and CogVideoX appear in \cref{fig:unified_appendix}.}
    \label{fig:presence_by_k}
    \vspace{-3mm}
\end{figure*}

\section{Experiments}
\label{sec:exp}

\subsection{Object Presence}
\paragraph{Backbones and baselines.}
We evaluate \method{} on LTX-2~\cite{hacohen2026ltx2} (the 22B LTX-2.3 checkpoint),
Wan~2.1~\cite{wanteam2025wan}, and CogVideoX-5B~\cite{yang2025cogvideox},
comparing against the unmodified generators, Attend-and-Excite~\cite{chefer2023attendexcite},
and Best-of-N seed resampling, which generates $N{=}3$ un-steered videos with different seeds and keeps the one that the evaluation detectors (described below) score best, i.e., all requested objects detected, then the highest detection ratio. Since the selection uses the evaluation detectors, Best-of-N is an oracle upper bound on seed resampling. We set $N{=}3$ to roughly match the cost of \method{} (\cref{tab:main}). %
Since no official video implementation of Attend-and-Excite is available, we re-implement it for all backbones using the objective in
\cref{eq:attend_excite}. We additionally report Wan~2.2~\cite{wanvideo2025wan22},
LTX-2.5~\cite{lightricks2026ltx25}, and MiniMax-H3~\cite{minimax2026h3}
as reference models. %

\paragraph{Benchmark and metrics.}
We evaluate on $473$ multi-object prompts with $k\in\{2,\ldots,7\}$ requested objects. We report detection ratio (det-ratio), the fraction of requested objects detected, and all-object accuracy (det-all), the fraction of videos in which all requested objects are detected. In addition, we evaluate on the VBench \texttt{object\_class} dimension, $79$ single-object prompts that serve as a guardrail. For single-object prompts, det-ratio and all-object accuracy coincide.
In these two benchmarks we additionally report VBench quality metrics to verify that improved presence does not compromise visual quality.  Full benchmark and evaluation details are provided in \cref{sec:app:evaluation}. 

\paragraph{Detectors.}
Object presence is measured using a majority vote of three detectors: GRiT, Grounding-DINO, and OWLv2, over 16 sampled frames per video; ties are excluded (details in \cref{sec:exp:metrics}).

\paragraph{Setup.}
\cref{tab:hparams} details the steering setup across LTX-2, Wan~2.1, and CogVideoX.

\paragraph{Inference cost.}
\method{} steers only the objects not yet predicted to be rendered, so in practice it averages $30$, $23$, and $13$ latent updates per clip on LTX-2, Wan~2.1, and CogVideoX, vs. $55$, $53$, and $54$ for A\&E. We report per-clip wall-clock time, averaged over nine generated videos across $k\in\{1,3,6\}$ per method on a single H100 GPU. Overall, \method{} requires $2.43\times$, $2.68\times$, and $2.07\times$ the base generation time on LTX-2, Wan~2.1, and CogVideoX, compared with $2.50\times$, $2.54\times$, and $2.66\times$ for A\&E (\cref{tab:main}); it is thus on par with A\&E on LTX-2 and Wan~2.1, and cheaper on CogVideoX.

\begin{table*}[t]
  \centering
  \footnotesize
  \setlength{\tabcolsep}{4pt}
  \caption{\textbf{Multi-object presence.}
  Detection ratio and all-object accuracy as the number of requested objects $k$
  increases, with VBench quality (7-dimension mean) and imaging quality (MUSIQ).
  Each backbone is shown as the un-steered base, Best-of-N, A\&E, and \method{}
  (Ours). \textbf{Bold} denotes the best within each backbone block;
  \underline{underline} the best overall.
  \emph{Inference cost} reports the mean number of latent optimization updates per clip, the wall-clock
  time per generated video on a single H100 GPU, and its multiple of the backbone's base generation time}
  \label{tab:main}
  \begin{tabular}{@{}l r c ccccccc cc ccc@{}}
    \toprule
    & & Det. & \multicolumn{7}{c}{All-object accuracy (\%) \up}
    & \multicolumn{2}{c}{Quality (\%) \up}
    & \multicolumn{3}{c}{Inference cost} \\
    \cmidrule(lr){4-10} \cmidrule(lr){11-12} \cmidrule(lr){13-15}
    Model & Params & ratio (\%) \up
    & mean & $k$=2 & 3 & 4 & 5 & 6 & 7 & VBench & MUSIQ
    & Updates\,\down & Time\,(s)\,\down & $\times$base\,\down \\
    \midrule
    Wan~2.2 & 5\,B & 78.9 & 38.3 & 87.5 & 52.5 & 35.0 & 35.0 & 7.5 & 9.6 & 84.4 & 75.4 & -- & -- & -- \\
    LTX-2.5 & 22\,B & 62.1 & 17.5 & 45.0 & 16.3 & 17.5 & 11.3 & 8.7 & 5.5 & \allbest{86.9} & 73.6 & -- & -- & -- \\
    MiniMax-H3 & 33\,B & 84.6 & 49.7 & 90.0 & 55.0 & 33.8 & 48.7 & 36.2 & 32.9 & 81.5 & 64.8 & -- & -- & -- \\
    \cmidrule(l){1-15}
    Wan~2.1 & 1.3\,B & 73.4 & 27.1 & 80.6 & 31.5 & 24.3 & 22.7 & 1.6 & 2.1 & 79.9 & 75.5 & 0 & 26.2 & $1.00\times$ \\
    \quad + Best-of-N & & 81.7 & 40.4 & \blockbest{93.6} & 59.5 & 35.9 & 28.2 & 11.8 & 13.2 & 80.1 & \blockbest{75.5} & 0 & 78.5 & $3.00\times$ \\
    \quad + A\&E & & 79.8 & 39.2 & 83.1 & 52.1 & 47.6 & 27.5 & 11.1 & 13.7 & 79.2 & 71.8 & 53 & 66.4 & $2.54\times$ \\
    \quad + Ours & & \blockbest{89.3} & \blockbest{59.9} & 90.9 & \blockbest{72.4} & \blockbest{60.9} & \blockbest{54.9} & \blockbest{41.3} & \blockbest{39.3} & \blockbest{82.5} & 74.6 & 23 & 70.2 & $2.68\times$ \\
    \cmidrule(l){1-15}
    LTX-2 & 22\,B & 78.9 & 38.3 & 75.6 & 60.3 & 35.5 & 26.3 & 20.0 & 12.1 & 83.7 & 70.6 & 0 & 59.8 & $1.00\times$ \\
    \quad + Best-of-N & & 85.9 & 51.1 & 88.6 & 67.5 & 49.4 & 45.6 & 30.4 & 25.0 & \blockbest{84.3} & 70.7 & 0 & 179.4 & $3.00\times$ \\
    \quad + A\&E & & 78.3 & 37.8 & 74.7 & 52.6 & 33.3 & 27.8 & 25.7 & 12.5 & 84.1 & 70.5 & 55 & 149.2 & $2.50\times$ \\
    \quad + Ours & & \blockbest{\allbest{93.9}} & \blockbest{\allbest{75.5}} & \blockbest{\allbest{95.0}} & \blockbest{\allbest{86.7}} & \blockbest{\allbest{84.4}} & \blockbest{\allbest{68.9}} & \blockbest{\allbest{65.7}} & \blockbest{\allbest{52.3}} & 83.5 & \blockbest{71.3} & 30 & 145.5 & $2.43\times$ \\
    \cmidrule(l){1-15}
    CogVideoX-5B & 5\,B & 73.1 & 30.5 & 80.0 & 54.1 & 20.3 & 13.4 & 8.6 & 6.5 & \blockbest{79.5} & 66.6 & 0 & 135.0 & $1.00\times$ \\
    \quad + Best-of-N & & 82.2 & 43.9 & 92.4 & 73.4 & 40.5 & 27.0 & 21.1 & 8.8 & 79.3 & 67.4 & 0 & 404.9 & $3.00\times$ \\
    \quad + A\&E & & 87.3 & 56.1 & 92.4 & 69.7 & 52.9 & 49.3 & 43.3 & 28.8 & 79.1 & 65.6 & 54 & 358.5 & $2.66\times$ \\
    \quad + Ours & & \blockbest{91.2} & \blockbest{65.9} & \blockbest{94.9} & \blockbest{81.8} & \blockbest{64.4} & \blockbest{61.1} & \blockbest{57.1} & \blockbest{36.2} & 78.4 & \blockbest{70.6} & 13 & 278.9 & $2.07\times$ \\
    \bottomrule
  \end{tabular}
\end{table*}

\paragraph{Multi-object presence.}
\Cref{tab:main} shows that \method{} improves all-object accuracy on all three backbones. \cref{fig:presence_by_k}(a) and \cref{fig:unified_appendix} show it visually for LTX-2, Wan~2.1 and CogVideoX respectively.
On Wan~2.1 it increases from $27.1\%$ to $59.9\%$, compared with
$39.2\%$ for A\&E and $40.4\%$ for Best-of-N; on LTX-2 from $38.3\%$ to
$75.5\%$ (vs $37.8\%$ and $51.1\%$); and on CogVideoX from $30.5\%$ to $65.9\%$ (vs $56.1\%$ and $43.9\%$). The advantage widens with $k$: at $k=7$ \method{} reaches $39.3\%$, $52.3\%$, and $36.2\%$, versus $13.7\%$, $12.5\%$, and $28.8\%$ for A\&E, while largely preserving VBench quality.

\paragraph{VBench single-object presence.}
The VBench \texttt{object\_class} single-object benchmark serves as a guardrail with a single requested object the backbones already succeed on most clips, and the goal is to verify that steering does not harm this easy regime. As shown in \cref{tab:singleobj} (\cref{sec:suppl:ablation}), \method{} in fact improves presence on LTX-2:
three-detector stays at $96.2\%$ (same as the baseline with the corrected detector),
and GRiT improves ($88.6\%\!\to\!89.9\%$) and on CogVideoX ($92.3\%\!\to\!96.2\%$ and
$81.0\%\!\to\!82.3\%$). On Wan~2.1 it is on par under the official GRiT
metric ($81.0\%$ for all three arms), with a small degradation under the
three-detector majority ($92.2\%\!\to\!88.3\%$). A\&E, in contrast, lowers
presence on every backbone (tying GRiT on Wan~2.1). \Cref{fig:singleobj} in \cref{sec:suppl:ablation} shows an example.

\begin{figure*}[t]
  \centering
  \includegraphics[width=\textwidth]{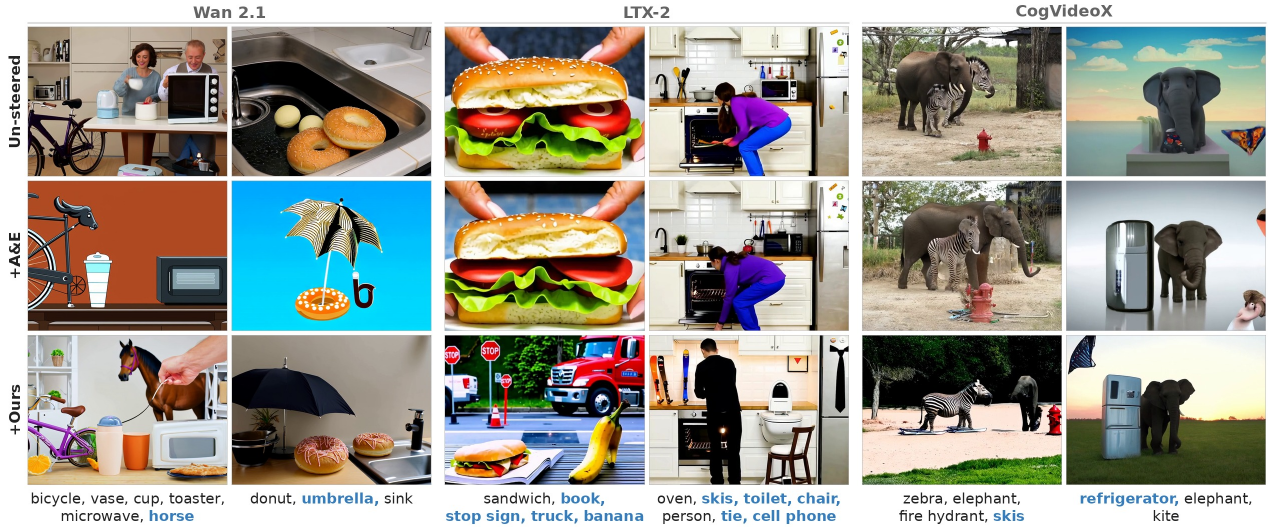}
  \vspace{-2mm}
  \caption{\textbf{Multi-object steering.}
  Qualitative results on Wan~2.1, LTX-2, and CogVideoX. Each column is one prompt, and its
  requested objects are listed below it; \textcolor{wacvblue}{\textbf{blue}}
  marks objects neglected by the un-steered backbone. The three videos of a prompt share the seed.
  \method{} recovers neglected objects while keeping the objects that the backbone already rendered.}
  \label{fig:compare}
  \vspace{-2mm}
\end{figure*}

\paragraph{Conflict-aware gradient aggregation (CAGA) ablation.}
\cref{tab:caga_ablation} shows that by removing CAGA (gradient aggregation + unit-norm) we lower all-object accuracy on every
backbone. The gap becomes wider with the number of requested objects~$k$, showing that gradient conflicts become more severe as prompts become more crowded. Averaged over the hard $k{=}5$--$7$ subset it costs $7.8$, $10.8$, and $18.1$ points on Wan~2.1, LTX-2, and CogVideoX respectively, whereas at low $k$ ($2$--$4$) the two variants are within standard error. This comes at no measurable cost in VBench quality (the VBench mean is within $1.5$ points on every backbone).

\paragraph{Qualitative results.}
\Cref{fig:compare} illustrates how the methods differ when recovering neglected
objects. On Wan~2.1, for ``bicycle, vase, cup, toaster, microwave, horse,'' \method{}
adds the horse to the kitchen of the backbone, whereas A\&E turns the scene into an
illustration, and for ``donut, umbrella, sink'' it adds the umbrella over the donuts
and the sink. On LTX-2, for ``sandwich, book, stop sign, truck, banana,'' both
baselines render only a close-up of the sandwich, whereas \method{} renders all five
objects, and for the seven-object kitchen prompt it adds the five missing objects.
On CogVideoX, \method{} adds the skis and the refrigerator.

\paragraph{User study.}
To confirm the improvement is visible to people, we ran a paired human-preference
study on LTX-2 and Wan~2.1, comparing \method{} against Attend-and-Excite on object
presence, photorealism, and motion smoothness (full protocol in
\cref{sec:suppl:userstudy}). \method{} is preferred on every question and on both
backbones. The largest single margin is object presence on Wan~2.1 ($73.9\%$ of
decisive votes), the quantity the method optimizes directly; photorealism and motion
smoothness --- included as quality guardrails --- also come out in our favor on both
backbones ($58$--$67\%$). These per-question preferences are shown for LTX-2 in
\cref{fig:presence_by_k}(b) and for Wan~2.1 in \cref{fig:unified_appendix}.

\subsection{Temporal Control}
\label{sec:exp:temporal}

We next ask whether the same learned objective can control \emph{when} an
object is rendered. This setting also subsumes persistent object presence, in which the object is requested in every frame.

The same shared head already yields frame-level logits at no extra cost.
Rather than pooling over space and time, we pool only spatially---keeping the
temporal axis---to read out frame-level rendering logits
$\boldsymbol{\ell}_i=(\ell_{i,1},\ldots,\ell_{i,f})$, where $\ell_{i,u}$
predicts whether object $O_i$ is rendered at frame $u$. Because the linear
$1{\times}1{\times}1$ head commutes with the frame mean, the video-level logit is
exactly their temporal average, $s_i=\mathrm{mean}_u\,\ell_{i,u}$; the frame-level
logits thus add no parameters and reuse the same self-generated videos and
detector labels as the video-level predictor, requiring no additional
supervision.

\paragraph{Temporal loss.}
Let $c_i(u)\in\{0,1\}$ specify whether $O_i$ should be rendered at frame $u$.
To keep longer intervals from dominating, we balance the two states:
\begin{equation}
\begin{aligned}
\mathcal{L}^{\mathrm{temp}}_i(\vz_t;c_i)
=\frac{1}{2}\Big[
&\mathbb{E}_{u:\,c_i(u)=1}
  \operatorname{BCE_H}(\ell_{i,u},1) \\
+{}&
\mathbb{E}_{u:\,c_i(u)=0}
  \operatorname{BCE_H}(\ell_{i,u},0)
\Big].
\end{aligned}
\label{eq:temporal_loss}
\end{equation}

Here $\operatorname{BCE_H}(\ell,y)=h_m\big(\operatorname{BCE}(\ell,y)\big)$ applies the Huber function $h_m(b)=\tfrac{1}{2}b^{2}$ for $b\le m$ and $m\big(b-\tfrac{m}{2}\big)$ otherwise to the binary cross-entropy $b$ between a frame logit $\ell$ and its requested state $y\in\{0,1\}$. Its gradient with respect to the logit, $\min(b,m)\,(\sigma(\ell)-y)$, grows with the error faster than plain BCE and is capped at $m$, which bounds the per-frame steering force. We use $m=2$ throughout; \cref{fig:bce_huber} in \cref{sec:suppl:hparams} plots this gradient, and \cref{sec:app:steering} details the regularization and optimization steps.

\paragraph{Baseline.}
We compare against TempoControl~\cite{tempocontrol}, which controls when objects
are rendered using hand-crafted objectives over their temporal cross-attention
patterns. It is a natural baseline here because both persistent presence and
time-varying appearance can be expressed directly as temporal control signals,
allowing a direct comparison between its predefined attention objective and our
learned frame-level objective. We re-implement TempoControl on LTX-2 and adjust its configuration.

\paragraph{Persistent two-object presence.}
We first consider VBench Multiple Objects (full results in \cref{tab:temporal_multi}, \cref{sec:app:temporal_results}), requiring
both requested objects throughout the video, i.e., $c_i(u)=1$ for all frames.
Compared with TempoControl, \method{} increases three-detector presence from
$79.1\%$ to $91.3\%$, t-IoU from $82.0\%$ to $92.8\%$, and R@0.5 from
$83.5\%$ to $92.7\%$, at slightly lower VBench quality.

\paragraph{Time-varying control.}
We next evaluate explicit changes in object presence
(\cref{tab:temporal}): a single object entering at a prescribed time, and a
static+dynamic setting where one object remains present while another enters
partway through. We report frame-wise precision, recall, specificity, and F1,
together with t-IoU and R@0.5 for temporal alignment.

For single-object control, \method{} improves t-IoU from $41.8\%$ to $45.0\%$
and R@0.5 from $37.5\%$ to $45.0\%$ over TempoControl, while increasing
specificity from $35.7\%$ to $52.6\%$. For static+dynamic control, t-IoU
improves from $70.8\%$ to $74.5\%$ and R@0.5 from $81.7\%$ to $85.4\%$,
with the best F1 of $85.5\%$. VBench quality is preserved in both settings.
\Cref{fig:temporal_beat_tempo} in \cref{sec:suppl:qual} shows qualitative examples.

Together, these results show that the learned objective supports both persistent
object presence and precise control over when objects are rendered.

\section{Limitations}

Our approach has three main limitations. (i) We currently steer using
cross-attention, which provides a natural connection between prompt tokens and
the evolving video, but other internal signals such as self-attention may carry
complementary information and could further improve steering. (ii) Learned
steering objectives are more sensitive to distribution shift than simple
hand-crafted objectives: aggressive optimization can move the latent away from
the trajectories seen during training, where the learned gradients may become
less reliable. For object presence, the gradient normalization and conflict-aware aggregation
keep each update close to the sampler's own trajectory; for temporal control we
add an explicit trust-region penalty (\cref{sec:app:optimization}).
(iii) Both our self-generated supervision and evaluation rely on imperfect
object detectors. We reduce dependence on any individual detector using a
three-detector majority vote, but automatic evaluation of object
presence remains challenging.

\section{Conclusion}

We presented \method{}, a learned inference-time steering approach for improving
object presence in text-to-video generation, replacing hand-crafted attention
objectives with a predictor of whether a requested object will be rendered. We
showed how to train this predictor efficiently using self-generated trajectories
and automatic supervision, without external video annotations. We further
addressed the challenge of jointly steering multiple objects with a
conflict-aware gradient aggregation rule. Finally, the same
learned principle extends from controlling \emph{whether} an
object is rendered to \emph{when} it is rendered.

\section*{Acknowledgments}
This research was supported by the Israel Science Foundation (ISF), grant no.\ 208075.

{
    \small
    \bibliographystyle{ieeenat_fullname}
    \bibliography{refs}

\begin{thebibliography}{56}
\providecommand{\natexlab}[1]{#1}
\providecommand{\url}[1]{\texttt{#1}}
\expandafter\ifx\csname urlstyle\endcsname\relax
  \providecommand{\doi}[1]{doi: #1}\else
  \providecommand{\doi}{doi: \begingroup \urlstyle{rm}\Url}\fi

\bibitem[Bai et~al.(2025)Bai, Chen, Liu, Wang, Ge, Song, Dang, Wang, Wang, Tang, et~al.]{bai2025qwen25vl}
Shuai Bai, Keqin Chen, Xuejing Liu, Jialin Wang, Wenbin Ge, Sibo Song, Kai Dang, Peng Wang, Shijie Wang, Jun Tang, et~al.
\newblock {Qwen2.5-VL} technical report.
\newblock \emph{arXiv preprint arXiv:2502.13923}, 2025.

\bibitem[Bansal et~al.(2023)Bansal, Chu, Schwarzschild, Sengupta, Goldblum, Geiping, and Goldstein]{bansal2023universal}
Arpit Bansal, Hong-Min Chu, Avi Schwarzschild, Soumyadip Sengupta, Micah Goldblum, Jonas Geiping, and Tom Goldstein.
\newblock Universal guidance for diffusion models.
\newblock In \emph{Proceedings of the IEEE/CVF Conference on Computer Vision and Pattern Recognition (CVPR) Workshops}, pages 843--852, 2023.

\bibitem[Betti et~al.(2024)Betti, Baraldi, Baraldi, Cucchiara, and Sebe]{betti2024head}
Federico Betti, Lorenzo Baraldi, Lorenzo Baraldi, Rita Cucchiara, and Nicu Sebe.
\newblock Optimizing resource consumption in diffusion models through hallucination early detection.
\newblock In \emph{Proceedings of the European Conference on Computer Vision (ECCV) Workshops}, pages 312--326. Springer, 2024.

\bibitem[Betti et~al.(2026)Betti, Baraldi, Baraldi, Cucchiara, and Sebe]{betti2026head}
Federico Betti, Lorenzo Baraldi, Lorenzo Baraldi, Rita Cucchiara, and Nicu Sebe.
\newblock Hallucination early detection in diffusion models.
\newblock \emph{International Journal of Computer Vision}, 134\penalty0 (1):\penalty0 35, 2026.

\bibitem[Blattmann et~al.(2023)Blattmann, Rombach, Ling, Dockhorn, Kim, Fidler, and Kreis]{blattmann2023videoldm}
Andreas Blattmann, Robin Rombach, Huan Ling, Tim Dockhorn, Seung~Wook Kim, Sanja Fidler, and Karsten Kreis.
\newblock Align your latents: High-resolution video synthesis with latent diffusion models.
\newblock In \emph{Proceedings of the IEEE/CVF Conference on Computer Vision and Pattern Recognition (CVPR)}, pages 22563--22575, 2023.

\bibitem[Chefer et~al.(2022)Chefer, Schwartz, and Wolf]{chefer2022optimizing}
Hila Chefer, Idan Schwartz, and Lior Wolf.
\newblock Optimizing relevance maps of vision transformers improves robustness.
\newblock In \emph{Advances in Neural Information Processing Systems}, pages 33618--33632, 2022.

\bibitem[Chefer et~al.(2023)Chefer, Alaluf, Vinker, Wolf, and Cohen-Or]{chefer2023attendexcite}
Hila Chefer, Yuval Alaluf, Yael Vinker, Lior Wolf, and Daniel Cohen-Or.
\newblock {Attend-and-Excite}: Attention-based semantic guidance for text-to-image diffusion models.
\newblock \emph{ACM Transactions on Graphics}, 42\penalty0 (4):\penalty0 148:1--148:10, 2023.

\bibitem[Dhariwal and Nichol(2021)]{dhariwal2021diffusion}
Prafulla Dhariwal and Alexander Nichol.
\newblock Diffusion models beat {GANs} on image synthesis.
\newblock In \emph{Advances in Neural Information Processing Systems}, pages 8780--8794, 2021.

\bibitem[Geng et~al.(2025)Geng, Herrmann, Hur, Cole, Zhang, Pfaff, Lopez-Guevara, Aytar, Rubinstein, Sun, Wang, Owens, and Sun]{geng2025motionprompting}
Daniel Geng, Charles Herrmann, Junhwa Hur, Forrester Cole, Serena Zhang, Tobias Pfaff, Tatiana Lopez-Guevara, Yusuf Aytar, Michael Rubinstein, Chen Sun, Oliver Wang, Andrew Owens, and Deqing Sun.
\newblock Motion prompting: Controlling video generation with motion trajectories.
\newblock In \emph{Proceedings of the IEEE/CVF Conference on Computer Vision and Pattern Recognition (CVPR)}, 2025.

\bibitem[HaCohen et~al.(2026)HaCohen, Brazowski, Chiprut, Bitterman, Kvochko, Berkowitz, Shalem, Lifschitz, Moshe, Porat, Richardson, Shiran, Chachy, Chetboun, Finkelson, Kupchick, Zabari, Guetta, Kotler, Bibi, Gordon, Panet, Benita, Armon, Kulikov, Inger, Shiftan, Melumian, and Farbman]{hacohen2026ltx2}
Yoav HaCohen, Benny Brazowski, Nisan Chiprut, Yaki Bitterman, Andrew Kvochko, Avishai Berkowitz, Daniel Shalem, Daphna Lifschitz, Dudu Moshe, Eitan Porat, Eitan Richardson, Guy Shiran, Itay Chachy, Jonathan Chetboun, Michael Finkelson, Michael Kupchick, Nir Zabari, Nitzan Guetta, Noa Kotler, Ofir Bibi, Ori Gordon, Poriya Panet, Roi Benita, Shahar Armon, Victor Kulikov, Yaron Inger, Yonatan Shiftan, Zeev Melumian, and Zeev Farbman.
\newblock {LTX-2}: Efficient joint audio-visual foundation model, 2026.

\bibitem[He et~al.(2016)He, Zhang, Ren, and Sun]{he2016deep}
Kaiming He, Xiangyu Zhang, Shaoqing Ren, and Jian Sun.
\newblock Deep residual learning for image recognition.
\newblock In \emph{Proceedings of the IEEE Conference on Computer Vision and Pattern Recognition (CVPR)}, pages 770--778, 2016.

\bibitem[Hertz et~al.(2023)Hertz, Mokady, Tenenbaum, Aberman, Pritch, and Cohen-Or]{hertz2022prompt}
Amir Hertz, Ron Mokady, Jay Tenenbaum, Kfir Aberman, Yael Pritch, and Daniel Cohen-Or.
\newblock Prompt-to-prompt image editing with cross-attention control.
\newblock In \emph{International Conference on Learning Representations}, 2023.

\bibitem[Ho et~al.(2020)Ho, Jain, and Abbeel]{ho2020denoising}
Jonathan Ho, Ajay Jain, and Pieter Abbeel.
\newblock Denoising diffusion probabilistic models.
\newblock In \emph{Advances in Neural Information Processing Systems}, pages 6840--6851, 2020.

\bibitem[Huang et~al.(2026)Huang, Cui, Ye, Dong, Chen, Xue, Yang, Huang, Hong, and Tang]{diffusionprobe2026}
Bukun Huang, Benlei Cui, Zhizeng Ye, Xuemei Dong, Tuo Chen, Hui Xue, Dingkang Yang, Longtao Huang, Haiwen Hong, and Jingqun Tang.
\newblock Diffusion probe: Generated image result prediction using {CNN} probes.
\newblock In \emph{Proceedings of the IEEE/CVF Conference on Computer Vision and Pattern Recognition (CVPR)}, pages 35926--35935, 2026.

\bibitem[Huang et~al.(2024)Huang, He, Yu, Zhang, Si, Jiang, Zhang, Wu, Jin, Chanpaisit, Wang, Chen, Wang, Lin, Qiao, and Liu]{huang2024vbench}
Ziqi Huang, Yinan He, Jiashuo Yu, Fan Zhang, Chenyang Si, Yuming Jiang, Yuanhan Zhang, Tianxing Wu, Qingyang Jin, Nattapol Chanpaisit, Yaohui Wang, Xinyuan Chen, Limin Wang, Dahua Lin, Yu Qiao, and Ziwei Liu.
\newblock {VBench}: Comprehensive benchmark suite for video generative models.
\newblock In \emph{Proceedings of the IEEE/CVF Conference on Computer Vision and Pattern Recognition (CVPR)}, pages 21807--21818, 2024.

\bibitem[Jain et~al.(2024)Jain, Nasery, Vineet, and Behl]{jain2024peekaboo}
Yash Jain, Anshul Nasery, Vibhav Vineet, and Harkirat Behl.
\newblock {PEEKABOO}: Interactive video generation via masked-diffusion.
\newblock In \emph{Proceedings of the IEEE/CVF Conference on Computer Vision and Pattern Recognition (CVPR)}, pages 8079--8088, 2024.

\bibitem[Ke et~al.(2021)Ke, Wang, Wang, Milanfar, and Yang]{ke2021musiq}
Junjie Ke, Qifei Wang, Yilin Wang, Peyman Milanfar, and Feng Yang.
\newblock {MUSIQ}: Multi-scale image quality transformer.
\newblock In \emph{Proceedings of the IEEE/CVF International Conference on Computer Vision (ICCV)}, pages 5148--5157, 2021.

\bibitem[Li et~al.(2023)Li, Keuper, Zhang, and Khoreva]{li2023divide}
Yumeng Li, Margret Keuper, Dan Zhang, and Anna Khoreva.
\newblock Divide \& bind your attention for improved generative semantic nursing.
\newblock In \emph{Proceedings of the British Machine Vision Conference (BMVC)}, 2023.

\bibitem[Lian et~al.(2024)Lian, Shi, Yala, Darrell, and Li]{lian2024lvd}
Long Lian, Baifeng Shi, Adam Yala, Trevor Darrell, and Boyi Li.
\newblock {LLM}-grounded video diffusion models.
\newblock In \emph{International Conference on Learning Representations}, 2024.

\bibitem[{Lightricks}(2026)]{lightricks2026ltx25}
{Lightricks}.
\newblock {LTX-2.5}: Video, audio \& world simulation.
\newblock \url{https://huggingface.co/Lightricks/LTX-2.5}, 2026.
\newblock Official model release, August 2026.

\bibitem[Lin et~al.(2014)Lin, Maire, Belongie, Hays, Perona, Ramanan, Doll{\'a}r, and Zitnick]{lin2014coco}
Tsung-Yi Lin, Michael Maire, Serge Belongie, James Hays, Pietro Perona, Deva Ramanan, Piotr Doll{\'a}r, and C.~Lawrence Zitnick.
\newblock Microsoft {COCO}: Common objects in context.
\newblock In \emph{European Conference on Computer Vision}, pages 740--755, 2014.

\bibitem[Liu et~al.(2021)Liu, Liu, Jin, Stone, and Liu]{liu2021cagrad}
Bo Liu, Xingchao Liu, Xiaojie Jin, Peter Stone, and Qiang Liu.
\newblock Conflict-averse gradient descent for multi-task learning.
\newblock In \emph{Advances in Neural Information Processing Systems}, pages 18878--18890, 2021.

\bibitem[Liu et~al.(2024)Liu, Zeng, Ren, Li, Zhang, Yang, Jiang, Li, Yang, Su, Zhu, and Zhang]{liu2024groundingdino}
Shilong Liu, Zhaoyang Zeng, Tianhe Ren, Feng Li, Hao Zhang, Jie Yang, Qing Jiang, Chunyuan Li, Jianwei Yang, Hang Su, Jun Zhu, and Lei Zhang.
\newblock {Grounding DINO}: Marrying {DINO} with grounded pre-training for open-set object detection.
\newblock In \emph{European Conference on Computer Vision (ECCV)}, 2024.

\bibitem[Luo et~al.(2024)Luo, Darrell, Wang, Goldman, and Holynski]{luo2024readout}
Grace Luo, Trevor Darrell, Oliver Wang, Dan~B Goldman, and Aleksander Holynski.
\newblock Readout guidance: Learning control from diffusion features.
\newblock In \emph{Proceedings of the IEEE/CVF Conference on Computer Vision and Pattern Recognition (CVPR)}, pages 8217--8227, 2024.

\bibitem[Mari~Oriyad et~al.(2025)Mari~Oriyad, Banayeeanzade, Abbasi, Rohban, and Soleymani~Baghshah]{marioriyad2025}
Arash Mari~Oriyad, Mohammadali Banayeeanzade, Reza Abbasi, Mohammad~Hossein Rohban, and Mahdieh Soleymani~Baghshah.
\newblock Attention overlap is responsible for the entity missing problem in text-to-image diffusion models!
\newblock \emph{Transactions on Machine Learning Research}, 2025.

\bibitem[Meral et~al.(2024)Meral, Simsar, Tombari, and Yanardag]{meral2024conform}
Tuna Han~Salih Meral, Enis Simsar, Federico Tombari, and Pinar Yanardag.
\newblock {CONFORM}: Contrast is all you need for high-fidelity text-to-image diffusion models.
\newblock In \emph{Proceedings of the IEEE/CVF Conference on Computer Vision and Pattern Recognition (CVPR)}, pages 9005--9014, 2024.

\bibitem[Minderer et~al.(2023)Minderer, Gritsenko, and Houlsby]{minderer2023owlv2}
Matthias Minderer, Alexey Gritsenko, and Neil Houlsby.
\newblock Scaling open-vocabulary object detection.
\newblock In \emph{Advances in Neural Information Processing Systems}, 2023.

\bibitem[{MiniMax}(2026)]{minimax2026h3}
{MiniMax}.
\newblock {MiniMax H3}.
\newblock \url{https://huggingface.co/MiniMaxAI/MiniMax-H3}, 2026.
\newblock Official model release, August 2026.

\bibitem[Mousai et~al.(2026)Mousai, Yiflach, Davidovitch, Mualem, Kraus, Barad, Kreis, and Chechik]{mousai2026fkremask}
Oded Mousai, Sapir~Esther Yiflach, Barak~Itzchak Davidovitch, Yuval Mualem, Sarit Kraus, Hannah-Noa Barad, Karsten Kreis, and Gal Chechik.
\newblock Multi-property {Feynman-Kac} steering of crystal diffusion models.
\newblock \emph{arXiv preprint arXiv:2610.24828}, 2026.

\bibitem[Peebles and Xie(2023)]{peebles2023dit}
William Peebles and Saining Xie.
\newblock Scalable diffusion models with transformers.
\newblock In \emph{Proceedings of the IEEE/CVF International Conference on Computer Vision (ICCV)}, pages 4195--4205, 2023.

\bibitem[Qu et~al.(2026)Qu, Wang, Zheng, Wang, Nie, and Chua]{qu2026ttom}
Leigang Qu, Ziyang Wang, Na Zheng, Wenjie Wang, Liqiang Nie, and Tat-Seng Chua.
\newblock {TTOM}: Test-time optimization and memorization for compositional video generation.
\newblock In \emph{International Conference on Learning Representations}, 2026.

\bibitem[Rahamim et~al.(2027)Rahamim, Samuel, Schwartz, and Chechik]{rahamim2027reccar}
Ohad Rahamim, Dvir Samuel, Idan Schwartz, and Gal Chechik.
\newblock All modalities are equal, but video is more equal: Closing the cross-attention gap in joint video generation.
\newblock In \emph{Proceedings of the IEEE/CVF Winter Conference on Applications of Computer Vision (WACV)}, 2027.

\bibitem[Rassin et~al.(2023)Rassin, Hirsch, Glickman, Ravfogel, Goldberg, and Chechik]{rassin2023syngen}
Royi Rassin, Eran Hirsch, Daniel Glickman, Shauli Ravfogel, Yoav Goldberg, and Gal Chechik.
\newblock Linguistic binding in diffusion models: Enhancing attribute correspondence through attention map alignment.
\newblock In \emph{Advances in Neural Information Processing Systems}, pages 3536--3559, 2023.

\bibitem[Rombach et~al.(2022)Rombach, Blattmann, Lorenz, Esser, and Ommer]{rombach2022highresolution}
Robin Rombach, Andreas Blattmann, Dominik Lorenz, Patrick Esser, and Bj{\"o}rn Ommer.
\newblock High-resolution image synthesis with latent diffusion models.
\newblock In \emph{Proceedings of the IEEE/CVF Conference on Computer Vision and Pattern Recognition (CVPR)}, pages 10684--10695, 2022.

\bibitem[Schiber et~al.(2026)Schiber, Lindenbaum, and Schwartz]{tempocontrol}
Shira Schiber, Ofir Lindenbaum, and Idan Schwartz.
\newblock {TempoControl}: Temporal attention guidance for text-to-video models.
\newblock In \emph{Proceedings of the IEEE/CVF Conference on Computer Vision and Pattern Recognition (CVPR)}, pages 36670--36679, 2026.

\bibitem[Schwartz et~al.(2017)Schwartz, Schwing, and Hazan]{schwartz2017high}
Idan Schwartz, Alexander~G. Schwing, and Tamir Hazan.
\newblock High-order attention models for visual question answering.
\newblock In \emph{Advances in Neural Information Processing Systems}, 2017.

\bibitem[Schwartz et~al.(2019)Schwartz, Yu, Hazan, and Schwing]{schwartz2019factor}
Idan Schwartz, Seunghak Yu, Tamir Hazan, and Alexander~G. Schwing.
\newblock Factor graph attention.
\newblock In \emph{Proceedings of the IEEE/CVF Conference on Computer Vision and Pattern Recognition (CVPR)}, 2019.

\bibitem[Schwartz et~al.(2023)Schwartz, Sn{\ae}bjarnarson, Chefer, Cotterell, Belongie, Wolf, and Benaim]{schwartz2023discriminative}
Idan Schwartz, V{\'e}steinn Sn{\ae}bjarnarson, Hila Chefer, Ryan Cotterell, Serge Belongie, Lior Wolf, and Sagie Benaim.
\newblock Discriminative class tokens for text-to-image diffusion models.
\newblock In \emph{Proceedings of the IEEE/CVF International Conference on Computer Vision (ICCV)}, pages 22725--22735, 2023.

\bibitem[Shaulov et~al.(2026)Shaulov, Shaar, Edenzon, Chechik, and Wolf]{shaulov2026cvg}
Ariel Shaulov, Eitan Shaar, Amit Edenzon, Gal Chechik, and Lior Wolf.
\newblock Compositional video generation via inference-time guidance, 2026.

\bibitem[Shpitzer et~al.(2026)Shpitzer, Weidenfeld, Chechik, and Schwartz]{shpitzer2026siso}
Yair Shpitzer, Aviv Weidenfeld, Gal Chechik, and Idan Schwartz.
\newblock Single image iterative subject-driven generation and editing.
\newblock In \emph{Proceedings of the IEEE/CVF Conference on Computer Vision and Pattern Recognition Workshops (CVPRW)}, 2026.

\bibitem[Singer et~al.(2023)Singer, Polyak, Hayes, Yin, An, Zhang, Hu, Yang, Ashual, Gafni, Parikh, Gupta, and Taigman]{singer2023makeavideo}
Uriel Singer, Adam Polyak, Thomas Hayes, Xi Yin, Jie An, Songyang Zhang, Qiyuan Hu, Harry Yang, Oron Ashual, Oran Gafni, Devi Parikh, Sonal Gupta, and Yaniv Taigman.
\newblock Make-a-video: Text-to-video generation without text-video data.
\newblock In \emph{International Conference on Learning Representations}, 2023.

\bibitem[Sun et~al.(2025)Sun, Huang, Liu, Wu, Xu, Li, and Liu]{sun2025t2vcompbench}
Kaiyue Sun, Kaiyi Huang, Xian Liu, Yue Wu, Zihan Xu, Zhenguo Li, and Xihui Liu.
\newblock {T2V-CompBench}: A comprehensive benchmark for compositional text-to-video generation.
\newblock In \emph{Proceedings of the IEEE/CVF Conference on Computer Vision and Pattern Recognition (CVPR)}, pages 8406--8416, 2025.

\bibitem[{Team Wan} et~al.(2025){Team Wan}, Wang, Ai, Wen, Mao, Xie, Chen, Yu, Zhao, Yang, Zeng, Wang, Zhang, Zhou, Wang, Chen, Zhu, Zhao, Yan, Huang, Feng, Zhang, Li, Wu, Chu, Feng, Zhang, Sun, Fang, Wang, Gui, Weng, Shen, Lin, Wang, Wang, Zhou, Wang, Shen, Yu, Shi, Huang, Xu, Kou, Lv, Li, Liu, Wang, Zhang, Huang, Li, Wu, Liu, Pan, Zheng, Hong, Shi, Feng, Jiang, Han, Wu, and Liu]{wanteam2025wan}
{Team Wan}, Ang Wang, Baole Ai, Bin Wen, Chaojie Mao, Chen-Wei Xie, Di Chen, Feiwu Yu, Haiming Zhao, Jianxiao Yang, Jianyuan Zeng, Jiayu Wang, Jingfeng Zhang, Jingren Zhou, Jinkai Wang, Jixuan Chen, Kai Zhu, Kang Zhao, Keyu Yan, Lianghua Huang, Mengyang Feng, Ningyi Zhang, Pandeng Li, Pingyu Wu, Ruihang Chu, Ruili Feng, Shiwei Zhang, Siyang Sun, Tao Fang, Tianxing Wang, Tianyi Gui, Tingyu Weng, Tong Shen, Wei Lin, Wei Wang, Wei Wang, Wenmeng Zhou, Wente Wang, Wenting Shen, Wenyuan Yu, Xianzhong Shi, Xiaoming Huang, Xin Xu, Yan Kou, Yangyu Lv, Yifei Li, Yijing Liu, Yiming Wang, Yingya Zhang, Yitong Huang, Yong Li, You Wu, Yu Liu, Yulin Pan, Yun Zheng, Yuntao Hong, Yupeng Shi, Yutong Feng, Zeyinzi Jiang, Zhen Han, Zhi-Fan Wu, and Ziyu Liu.
\newblock Wan: Open and advanced large-scale video generative models, 2025.

\bibitem[Tian et~al.(2024)Tian, Yang, Yang, Gao, Deng, Chen, Wang, Yu, Tao, Wan, Zhang, and Cui]{tian2024videotetris}
Ye Tian, Ling Yang, Haotian Yang, Yuan Gao, Yufan Deng, Jingmin Chen, Xintao Wang, Zhaochen Yu, Xin Tao, Pengfei Wan, Di Zhang, and Bin Cui.
\newblock {VideoTetris}: Towards compositional text-to-video generation.
\newblock In \emph{Advances in Neural Information Processing Systems}, pages 29489--29513, 2024.

\bibitem[{Wan Team}(2025)]{wanvideo2025wan22}
{Wan Team}.
\newblock {Wan2.2}.
\newblock \url{https://huggingface.co/Wan-AI/Wan2.2-T2V-A14B}, 2025.
\newblock Official model release, July 2025.

\bibitem[Wang et~al.(2024{\natexlab{a}})Wang, Chen, Liu, Chen, Lin, Han, and Ding]{wang2024yolov10}
Ao Wang, Hui Chen, Lihao Liu, Kai Chen, Zijia Lin, Jungong Han, and Guiguang Ding.
\newblock {YOLOv10}: Real-time end-to-end object detection.
\newblock In \emph{Advances in Neural Information Processing Systems}, 2024{\natexlab{a}}.

\bibitem[Wang et~al.(2024{\natexlab{b}})Wang, Zhang, Zou, Zeng, Wei, Yuan, and Li]{wang2024boximator}
Jiawei Wang, Yuchen Zhang, Jiaxin Zou, Yan Zeng, Guoqiang Wei, Liping Yuan, and Hang Li.
\newblock Boximator: Generating rich and controllable motions for video synthesis.
\newblock In \emph{Proceedings of the 41st International Conference on Machine Learning}, pages 52274--52289. PMLR, 2024{\natexlab{b}}.

\bibitem[Wu et~al.(2022)Wu, Wang, Yang, Gan, Liu, Yuan, and Wang]{wu2022grit}
Jialian Wu, Jianfeng Wang, Zhengyuan Yang, Zhe Gan, Zicheng Liu, Junsong Yuan, and Lijuan Wang.
\newblock {GRiT}: A generative region-to-text transformer for object understanding, 2022.

\bibitem[Wu et~al.(2025)Wu, Siarohin, Menapace, Skorokhodov, Fang, Chordia, Gilitschenski, and Tulyakov]{wu2025mind}
Ziyi Wu, Aliaksandr Siarohin, Willi Menapace, Ivan Skorokhodov, Yuwei Fang, Varnith Chordia, Igor Gilitschenski, and Sergey Tulyakov.
\newblock Mind the time: Temporally-controlled multi-event video generation.
\newblock In \emph{Proceedings of the IEEE/CVF Conference on Computer Vision and Pattern Recognition (CVPR)}, pages 23989--24000, 2025.

\bibitem[Xu et~al.(2026)Xu, Song, Cai, and Zhang]{xu2026switchcraft}
Qianxun Xu, Chenxi Song, Yujun Cai, and Chi Zhang.
\newblock {SwitchCraft}: Training-free multi-event video generation with attention controls.
\newblock In \emph{Proceedings of the IEEE/CVF Conference on Computer Vision and Pattern Recognition (CVPR)}, pages 29136--29145, 2026.

\bibitem[Yang and Wang(2024)]{yang2024vico}
Xingyi Yang and Xinchao Wang.
\newblock Compositional video generation as flow equalization.
\newblock \emph{arXiv preprint arXiv:2407.06182}, 2024.

\bibitem[Yang et~al.(2025)Yang, Teng, Zheng, Ding, Huang, Xu, Yang, Hong, Zhang, Feng, Yin, Zhang, Wang, Cheng, Xu, Gu, Dong, and Tang]{yang2025cogvideox}
Zhuoyi Yang, Jiayan Teng, Wendi Zheng, Ming Ding, Shiyu Huang, Jiazheng Xu, Yuanming Yang, Wenyi Hong, Xiaohan Zhang, Guanyu Feng, Da Yin, Yuxuan Zhang, Weihan Wang, Yean Cheng, Bin Xu, Xiaotao Gu, Yuxiao Dong, and Jie Tang.
\newblock {CogVideoX}: Text-to-video diffusion models with an expert transformer.
\newblock In \emph{International Conference on Learning Representations}, 2025.

\bibitem[Yariv et~al.(2024)Yariv, Gat, Benaim, Wolf, Schwartz, and Adi]{yariv2024diverse}
Guy Yariv, Itai Gat, Sagie Benaim, Lior Wolf, Idan Schwartz, and Yossi Adi.
\newblock Diverse and aligned audio-to-video generation via text-to-video model adaptation.
\newblock In \emph{Proceedings of the AAAI Conference on Artificial Intelligence}, pages 6639--6647, 2024.

\bibitem[Yiflach et~al.(2026)Yiflach, Atzmon, and Chechik]{learn2steer}
Sapir~Esther Yiflach, Yuval Atzmon, and Gal Chechik.
\newblock Data-driven loss functions for inference-time optimization in text-to-image.
\newblock In \emph{Proceedings of the IEEE/CVF Winter Conference on Applications of Computer Vision (WACV)}, pages 3525--3535, 2026.

\bibitem[Yu et~al.(2020)Yu, Kumar, Gupta, Levine, Hausman, and Finn]{yu2020pcgrad}
Tianhe Yu, Saurabh Kumar, Abhishek Gupta, Sergey Levine, Karol Hausman, and Chelsea Finn.
\newblock Gradient surgery for multi-task learning.
\newblock In \emph{Advances in Neural Information Processing Systems}, pages 5824--5836, 2020.

\bibitem[Zafar et~al.(2026)Zafar, Cohen, Wolf, and Schwartz]{zafar2026detection}
Oz Zafar, Yuval Cohen, Lior Wolf, and Idan Schwartz.
\newblock Detection-driven object count optimization for text-to-image diffusion models.
\newblock In \emph{Proceedings of the IEEE/CVF Winter Conference on Applications of Computer Vision (WACV)}, 2026.

\end{thebibliography}
}
\clearpage
\onecolumn
\begin{center}
  {\LARGE\bfseries Supplementary Material\par}
  \vspace{1em}
\end{center}
\begin{multicols}{2}
  \tableofcontents
\end{multicols}

\appendix
\addtocontents{toc}{\protect\setcounter{tocdepth}{4}}
\let\supploldsection\section
\renewcommand{\section}{\clearpage\supploldsection}

\section{\method{} Model}
\label{sec:app:method}

\subsection{Training Data}
    \paragraph{Prompts dataset.}
    Training examples are generated from templated multi-object prompts: we use a fixed pool of $88$ object nouns, comprising $15$ animals and $73$ everyday objects, together with $26$ environments (\cref{tab:prompt_pool}). The pool starts from the $79$ nouns of VBench's \texttt{object\_class} dimension --- the same COCO-derived vocabulary~\cite{lin2014coco} the presence benchmarks score against --- renaming two of them for disambiguation (\emph{keyboard}$\rightarrow$\emph{computer keyboard}, \emph{remote}$\rightarrow$\emph{tv remote}), and adds nine nouns that dimension does not cover: five animals (\emph{rabbit}, \emph{lion}, \emph{fox}, \emph{owl}, \emph{penguin}) and four objects (\emph{computer mouse}, \emph{guitar}, \emph{lantern}, \emph{telescope}).

\begin{table}[b]
  \small
  \caption{Object and environment pool for the templated prompts: $88$ nouns
  ($15$ animals, $73$ everyday objects) and $26$ environments.}
  \label{tab:prompt_pool}
  \begin{tabular}{@{}p{0.23\linewidth}p{0.71\linewidth}@{}}
    \toprule
    Animals (15) & bear, bird, cat, cow, dog, elephant, giraffe, horse, sheep, zebra, rabbit, lion, fox, owl, penguin \\
    \midrule
    Objects (73) & airplane, apple, backpack, banana, baseball bat, baseball glove, bed, bench, bicycle, boat, book, bottle, bowl, broccoli, bus, cake, car, carrot, cell phone, chair, clock, computer keyboard, computer mouse, couch, cup, dining table, donut, fire hydrant, fork, frisbee, hair drier, handbag, hot dog, kite, knife, laptop, microwave, motorcycle, orange, oven, parking meter, person, pizza, potted plant, refrigerator, sandwich, scissors, sink, skateboard, skis, snowboard, spoon, sports ball, stop sign, suitcase, surfboard, teddy bear, tennis racket, tie, toaster, toilet, toothbrush, traffic light, train, truck, tv, tv remote, umbrella, vase, wine glass, guitar, lantern, telescope \\
    \midrule
    Environments (26) & park, beach, forest, kitchen, library, rooftop, cave, snowy mountain, desert, garden, house, office, street, space station, living room, classroom, restaurant, stadium, airport, train station, market, farm, circus, festival, wedding, birthday party \\
    \bottomrule
  \end{tabular}
\end{table}
For each object count $k$, object sets are sampled uniformly without replacement under a fixed random seed.
Prompts enumerate the sampled objects and are optionally followed by an environment phrase.
Each object set is instantiated once without an environment and twice with randomly sampled environments, deduplicated by prompt string, and generated using $6$--$8$ random generation seeds on LTX-2 and one seed on Wan~2.1; after filtering, the Wan~2.1 data covers $9{,}261$ prompts.
We combine prompt lists covering different ranges of $k$, yielding training prompts with $k=1,\ldots,7$ requested objects.
The evaluation prompts are constructed separately and share no prompt text with the training set.

Because objects are sampled independently rather than according to natural co-occurrence statistics, these prompts frequently contain unusual object combinations.
This provides challenging supervision for object presence, rather than allowing the generator to rely primarily on familiar scene statistics.

    \paragraph{Data labeling.}
\label{par:data_labeling}
Individual open-vocabulary detectors are not reliable enough for this purpose, and
training is sensitive to label noise. We therefore label each frame by a majority
vote over three detectors, GRiT~\cite{wu2022grit}, Grounding-DINO~\cite{liu2024groundingdino}
and OWLv2~\cite{minderer2023owlv2}.
Cases without a reliable majority are discarded rather than assigned a binary
label. Training labels aggregate $16$ middle-sampled frames, and an object counts
as present when it is detected in at least $15\%$ of them (\cref{sec:app:evaluation}).

    \paragraph{Data filtering.}
The self-generated dataset is naturally imbalanced because some objects are almost always rendered while others are frequently neglected.
We therefore retain object classes only when their positive-negative ratio is no more extreme than $15\%$--$85\%$.
The validation set is additionally rebalanced so that held-out object words are sufficiently represented.

After generation and filtering, the LTX-2 dataset holds $28{,}336$
$(\text{clip},\text{object})$ examples over $15{,}907$ clips, split into $23{,}679$ training examples ($70.4\%$ positive across the whole set) and $4{,}657$ validation examples ($72.3\%$ positive). Wan~2.1 is built the same way and is smaller: $19{,}117$ training
examples ($7{,}732$ clips, $68.7\%$ positive) and $3{,}814$ validation examples
($1{,}529$ clips, $71.1\%$ positive). Splits are grouped
by prompt identity, so no prompt text appears in both training and validation.

    \paragraph{Feature extraction.}
    Each prompt is generated with the backbone's standard sampler, and the
cross-attention maps are captured at a fixed set of denoising steps. A clip
contributes one training example per (target object, captured step) pair,
labeled positive if the object was rendered and negative otherwise. We capture
every transformer layer and head, so one example is the object-specific tensor
$\attn_i(\vz_t)\in\mathbb{R}^{L\times H\times f\times h\times w}$ of
the main paper. We capture $t\in\{0,1,4,7,10,15,20\}$ on LTX-2 and
$t\in\{0,1,4,7,10,15\}$ on Wan~2.1 and CogVideoX. LTX-2 has $L{=}48$ layers, $H{=}32$ heads and a
$13{\times}16{\times}16$ latent grid, and Wan~2.1 has $L{=}30$, $H{=}12$ and a
$6{\times}30{\times}52$ grid for its $21$ frames; on both, the predictor consumes
this tensor in full. CogVideoX is built the same way; it has $L{=}42$ layers and
$H{=}48$ heads, and its predictor consumes $21$ of the layers.

\subsection{Training Implementation Details}
\label{subsec:app:impl}

\begin{figure*}[t]
  \centering
  \includegraphics[width=\textwidth]{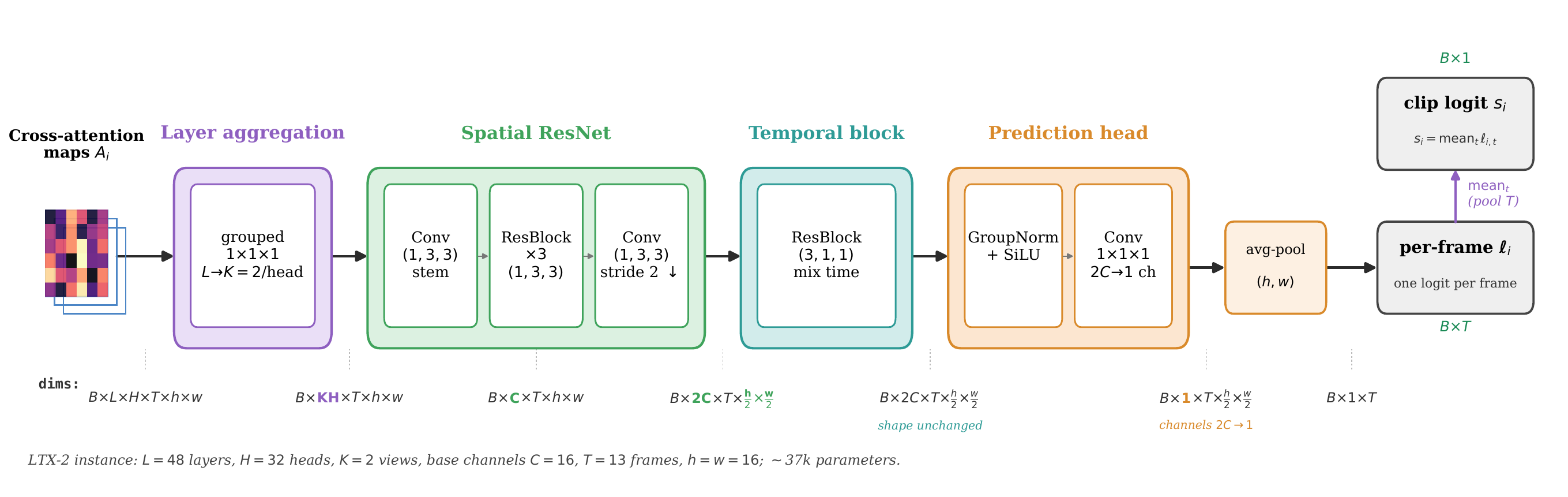}
  \caption{\textbf{\method{} architecture.} The per-(layer, head) cross-attention
  maps $A_i$ (plus the mean text key per head) enter at the left. A learned
  grouped $1{\times}1{\times}1$ convolution aggregates the $L$ layers into $K{=}2$
  views per head; a spatial ResNet (a stem and three $(1,3,3)$ residual blocks
  with downsampling) processes each frame independently; a final $(3,1,1)$
  residual block mixes across time. A single shared linear $1{\times}1{\times}1$ head
  (channels $2C{\to}1$) followed by a spatial pool over $(h,w)$ gives the per-frame
  logits $\boldsymbol{\ell}_i$ used for temporal control (\cref{sec:exp:temporal}); the
  clip logit is their temporal mean, $s_i=\mathrm{mean}_t\,\ell_{i,t}$ (equivalently a
  global pool over $(T,h,w)$, since the head is linear over the frame mean). Shapes are the
  LTX-2 instance ($L{=}48$ layers, $H{=}32$ heads, $13{\times}16{\times}16$
  latent grid); $\sim$37k parameters in total.}
  \label{fig:probe_arch}
\end{figure*}

\paragraph{Architecture.}
We train an independent \method{} model for each captured step $t$; a steered step without one uses the nearest predictor. The architecture is described in
\cref{sec:method:attnguide} of the main text and illustrated in
\cref{fig:probe_arch}.

\paragraph{Training hyper-parameters.}
The training hyper-parameters are listed in \cref{tab:train_hparams}. Both
early stopping and per-step model selection use the validation loss.

\begin{table}[t]
  \centering
  \footnotesize
  \caption{\method{} training hyper-parameters (one predictor per steering
  step). Values are shared by all backbones except where noted.}
  \label{tab:train_hparams}
  \begin{tabular}{@{}ll@{}}
    \toprule
    Optimizer & AdamW \\
    Learning rate & $3\times10^{-3}$ \\
    Weight decay & $1\times10^{-4}$ (LTX-2) \,/\, $1\times10^{-5}$ (Wan~2.1) \\
    LR schedule & cosine annealing (over max epochs) \\
    Batch size & $128$ \\
    Max epochs & $200$ (with early stopping) \\
    Early-stop patience & $50$ (LTX-2) \,/\, $40$ (Wan~2.1) epochs \\
    Normalization & instance norm (\texttt{InstanceNorm3d}) \\
    DropBlock (residual stack) & rate $0.3$, block size $3$ (\texttt{DropBlock3d}) \\
    Head dropout & $0.3$ (channel-wise \texttt{Dropout3d}) \\
    Label smoothing & $0.05$ (Wan~2.1 only) \\
    \bottomrule
  \end{tabular}
\end{table}

\paragraph{Loss.}
Residual class imbalance is handled by inverse-frequency weighting of the binary
cross-entropy objective over class groups. The resulting sample weights are
rescaled to have mean one, which preserves approximately the same overall loss
scale as the unweighted objective.
    
\paragraph{Training regularization.}
Despite its small parameter count, \method{} can overfit the generated training
data. We therefore combine instance normalization, dropout and DropBlock inside
the residual stack with early stopping. Because a separate predictor is trained
for every steering step, model selection is performed independently at each
step.

\section{Steering Details}
\label{sec:app:steering}

\subsection{Steering Regularization}
\label{sec:app:optimization}

Repeated steering can move the latent away from the generator's natural sampling
trajectory. For object presence, the gradient normalization and mutual-agreement
aggregation already keep the update close to that trajectory, so no explicit
regularizer is needed and all of its weights are set to zero. Temporal control
does not use that aggregation --- it steers only one or two objects --- so there
we add an explicit trust-region penalty,
\begin{equation}
\begin{aligned}
\mathcal{R}(\vz)
={}&
\lambda_{\mathrm{a}}
\left\|\vz-\vz_t^{(0)}\right\|_2^2 \\
&+
\lambda_{\mathrm{p}}
\left\|
\hat{\vx}_0(\vz)
-
\hat{\vx}_0(\vz_t^{(0)})
\right\|_2^2 \\
&+
\lambda_{\mathrm{m}}
\left(
\|\vz\|_2-\|\vz_t^{(0)}\|_2
\right)^2 \\
&+
\lambda_{\mathrm{KL}}\,
\mathrm{KL}\!\left(\operatorname{softmax}(\vz)\,\middle\|\,\operatorname{softmax}(\vz_t^{(0)})\right),
\end{aligned}
\label{eq:trust_region}
\end{equation}

where the first term limits displacement from the sampler's original latent at
the current denoising step; the second limits changes in the predicted denoised
sample $\hat{\vx}_0$, a constraint more directly tied to changes in the
generated content; the third preserves the latent magnitude expected by the
diffusion noise schedule; and the fourth keeps the softmax-normalized latent close to that of the sampler's latent. The generator and \method{} remain frozen throughout
optimization; gradients are propagated through \method{}, the attention
computation and the diffusion transformer to the latent. The weights we use are
listed in \cref{tab:hparams}.

\subsection{Optimization Steps}
For the temporal task, we steer at each of the first eleven denoising steps ($t\in\{0,\dots,10\}$); otherwise, the sampler reverts the steering. Instead of training a predictor for each of these steps, we use the predictor of the nearest captured step. Presence steering on Wan~2.1 and CogVideoX uses the same rule for steered steps without a predictor of their own.

In \cref{eq:temporal_loss}, each expectation is taken uniformly over the frames in the corresponding state; if one state is absent, only the other term is used.

\subsection{Steering Configuration}
\label{sec:suppl:hparams}
\Cref{tab:hparams} lists the steering hyper-parameters of every arm we report,
per backbone and task. \Cref{fig:bce_huber} plots the gradient of the capped
temporal loss.

\noindent
\begin{minipage}[t]{0.37\linewidth}
  \centering
  \captionof{figure}{\textbf{Gradient of the capped temporal loss $\operatorname{BCE_H}$ of
  \cref{eq:temporal_loss}.} Magnitude of $\partial\mathcal{L}/\partial\ell$
  against \method{}'s output for a single frame, where $\ell$ is that frame's
  rendering logit (\cref{sec:exp:temporal}). \textbf{left:} object is requested
  at this frame ($c{=}1$). \textbf{right:} object requested absent ($c{=}0$).
  It punishes harder than plain BCE when the prediction is far from the target
  without exploding the way the uncapped mirror
  $\tfrac{1}{2}\operatorname{BCE}^{2}$ does --- past the knee
  $\operatorname{BCE}{=}m$ it saturates at the ceiling $m$ --- and vanishes
  close to the target.}
  \label{fig:bce_huber}
  \includegraphics[width=\linewidth]{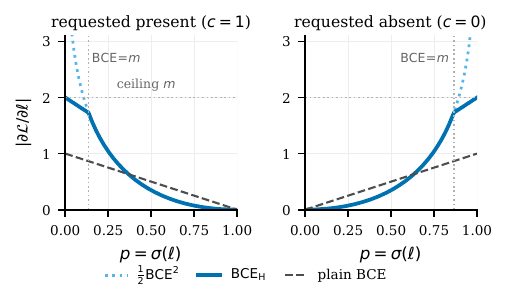}
\end{minipage}\hfill
\begin{minipage}[t]{0.60\linewidth}
  \centering\footnotesize
    \centering\footnotesize
  \setlength{\tabcolsep}{4pt}
  \captionof{table}{\textbf{Steering configuration.} Hyper-parameters of every steered
  arm we report, per backbone and task. All arms optimize the latent with a frozen steering model;
  $\lambda_{\mathrm{a}},\lambda_{\mathrm{p}},\lambda_{\mathrm{m}},\lambda_{\mathrm{KL}}$ are the
  trust-region weights of \cref{eq:trust_region}.}
  \label{tab:hparams}
  \resizebox{\linewidth}{!}{%
  \begin{tabular}{l ccc c}
    \toprule
    & \multicolumn{3}{c}{Presence} & Temporal \\
    \cmidrule(lr){2-4} \cmidrule(lr){5-5}
    & LTX-2 & Wan~2.1 & CogVideoX & LTX-2 \\
    \midrule
    Resolution                       & $512{\times}512$ & $832{\times}480$ & $720{\times}480$ & $512{\times}512$ \\
    Frames                           & 97 & 21 & 49 & 97 \\
    Denoising steps $T$              & 50 & 50 & 50 & 50 \\
    Guidance scale                   & 5.0 & 6.0 & 6.0 & 5.0 \\
    \midrule
    Steered steps $\mathcal{S}$      & $\{0,1,4,7,10,15\}$ & $\{1,\dots,15\}$ & $\{0,1,4,7,10,15\}$ & $\{0,\dots,10\}$ \\
    Iterations / step $N$            & 5 & 8 & 5 & 5 \\
    Step size $\eta$                 & 15 & 5 & 5 & 10 \\
    Gate $\tau$                      & 0.95 & 0.95 & 0.95 & -- \\
    All-satisfied fallback           & worst & zero & zero & -- \\
    Max updates $|\mathcal{S}|{\times}N$ & 30 & 120 & 30 & 55 \\
    \midrule
    Anchor $\lambda_{\mathrm{a}}$    & 0 & 0 & 0 & 1700 \\
    Prediction $\lambda_{\mathrm{p}}$& 0 & 0 & 0 & 0 \\
    Magnitude $\lambda_{\mathrm{m}}$ & 0 & 0 & 0 & 0.04 \\
    Softmax-KL $\lambda_{\mathrm{KL}}$ & 0 & 0 & 0 & 100 \\
    \midrule
    Supervision                      & clip logit & clip logit & clip logit & per-frame \\
    \bottomrule
  \end{tabular}}

\end{minipage}

\subsection{Full Temporal Results}
\label{sec:app:temporal_results}
\Cref{tab:temporal} reports every metric of the temporal-control and two-object experiments of \cref{sec:exp:temporal}.

\begin{table*}[h]
  \centering
\caption{\textbf{Temporal control and two-object presence.}
\textbf{Single object}: a requested object must enter at a prescribed
time and remain absent beforehand.
\textbf{Static + dynamic}: one object is present throughout the video
while a second enters partway through.
\textbf{Two objects}: VBench Multiple Objects with both objects required
throughout the video (3-det.\ is our 3-detector majority).
Frame-wise metrics measure presence and absence at each frame, while t-IoU and
R@0.5 measure alignment with the requested temporal interval. VBench reports
video quality. All values are in \%; \textbf{bold} denotes the best result in
each setting.}
  \label{tab:temporal}
  \label{tab:temporal_multi}
  \setlength{\tabcolsep}{3pt}
  \resizebox{\textwidth}{!}{%
  \begin{tabular}{@{}l cccc cc c | cccc cc c | cc cc c@{}}
    \toprule
    & \multicolumn{7}{c|}{Single object ($n{=}80$)}
    & \multicolumn{7}{c|}{Static + dynamic ($n{=}82$)}
    & \multicolumn{5}{c}{Two objects} \\
    \cmidrule(lr){2-8} \cmidrule(lr){9-15} \cmidrule(lr){16-20}

    & \multicolumn{4}{c}{Frame-wise \up}
    & \multicolumn{2}{c}{Temp.\ align.\ \up}
    & VBench \up
    & \multicolumn{4}{c}{Frame-wise \up}
    & \multicolumn{2}{c}{Temp.\ align.\ \up}
    & VBench \up
    & \multicolumn{2}{c}{Presence \up}
    & \multicolumn{2}{c}{Temp.\ align.\ \up}
    & VBench \up \\
    \cmidrule(lr){2-5} \cmidrule(lr){6-7}
    \cmidrule(lr){9-12} \cmidrule(lr){13-14}
    \cmidrule(lr){16-17} \cmidrule(lr){18-19}

    Model
    & Prec. & Rec. & Spec. & F1 & t-IoU & R@0.5 &
    & Prec. & Rec. & Spec. & F1 & t-IoU & R@0.5 &
    & 3-det. & YOLO & t-IoU & R@0.5 & \\
    \midrule

    LTX-2
      & 55.9 & 78.9 & 31.2 & 58.6 & 40.6 & 37.5 & 74.7
      & 86.4 & 77.1 & \blockbest{54.1} & 83.0 & 67.4 & 78.7 & 77.7
      & 78.1 & 56.8 & 80.7 & 82.9 & \blockbest{83.9} \\

    \quad + TempoControl
      & 57.7 & \blockbest{80.9} & 35.7 & 60.5 & 41.8 & 37.5 & 74.3
      & 87.1 & 80.5 & 53.7 & 83.1 & 70.8 & 81.7 & 78.6
      & 79.1 & 59.1 & 82.0 & 83.5 & 83.6 \\

    \quad + Ours
      & \blockbest{63.8} & 68.1 & \blockbest{52.6}
      & \blockbest{62.5} & \blockbest{45.0} & \blockbest{45.0}
      & \blockbest{75.1}
      & \blockbest{87.2} & \blockbest{84.1} & 52.2
      & \blockbest{85.5} & \blockbest{74.5} & \blockbest{85.4}
      & \blockbest{78.9}
      & \blockbest{91.3} & \blockbest{69.6} & \blockbest{92.8} & \blockbest{92.7} & 81.0 \\
    \bottomrule
  \end{tabular}%
  }
\end{table*}
\section{Evaluation}
\label{sec:app:evaluation}

\subsection{Benchmarks}
\label{sec:exp:benchmarks}
Our primary benchmark is a set of $473$ multi-object prompts spanning
$k\in\{2,\ldots,7\}$ requested objects: $80$ prompts for each of $k=2,\ldots,6$
and $73$ for $k=7$. We follow the VBench \texttt{multiple\_objects} dimension, use its $76$-class list and its prompt template (e.g., ``a X and a Y''), and extend it from two objects (our $k=2$ list) to seven objects. For each prompt, we randomly sample $k$ objects out of the $76$ classes under a pairwise-uniqueness constraint --- no two objects that co-occur in one prompt ever co-occur in any other prompt.
Because objects are sampled independently of
natural co-occurrence, the prompts frequently request unusual object
combinations; this makes object presence hard and prevents the generator from relying on familiar scene statistics. All backbones (LTX-2, Wan~2.1, CogVideoX) and arms (un-steered, Best-of-N, A\&E, and ours) are evaluated on the same $473$ prompts. Each prompt has its own random seed, shared by all arms; Best-of-N draws $2$ additional seeds per prompt.\\
The temporal-control benchmarks are TempoControl's own prompt lists, used unchanged.
\Cref{tab:benchmarks} summarizes the structure of all benchmarks.
\begin{table*}[t]
  \centering\footnotesize
  \setlength{\tabcolsep}{4pt}
  \caption{\textbf{Structure of each evaluation benchmark.}
  The temporal sets also carry a per-frame control signal.}
  \label{tab:benchmarks}
  \begin{tabular}{@{}l c c c l l@{}}
    \toprule
    Set & \makecell[c]{Size \\ {[\# prompts]}} & \makecell[c]{\# Obj./ \\ prompt} & \# Classes & Metrics & Origin \\
    \midrule
    
    \textbf{Presence} &&&&&\\
    Multi-objects 473 (\cref{tab:main}) &  473 &  2--7 &   76 & det-ratio, all-object acc. & \makecell[l]{extension of VBench\\{\texttt{multiple\_objects}}} \\
    Single-object (\cref{tab:singleobj})    &   79  &     1 &   79 & det-ratio (= all-object acc.) & VBench \texttt{object\_class} \\
    \midrule
    \textbf{Temporal} &&&&&\\
    Single object (\cref{tab:temporal} left)     &   80  &     1 &   80 & frame-wise P/R/Spec./F1, t-IoU, R@0.5 & TempoControl \\
    Static+dynamic (\cref{tab:temporal} right) &   82  &     2 &   80 & frame-wise P/R/Spec./F1, t-IoU, R@0.5 & TempoControl \\
    Two objects present (\cref{tab:temporal_multi}) &   82  &     2 &   76 & 3-det., YOLOv10, t-IoU, R@0.5 & VBench \texttt{multiple\_objects}\\

    \bottomrule
  \end{tabular}
\end{table*}

\subsection{Metrics Preliminaries}
\label{sec:exp:metrics}
\paragraph{Object detectors.}
Object detectors provide the raw signal --- whether an object is present in a
frame or a clip --- on which every reported metric is built, so a metric's
value depends on the detector used.\\
Each benchmark comes with its own detector, and for a fair comparison we always report metrics under the benchmark's native model. Specifically, for \emph{two-object presence} we report YOLOv10~\cite{wang2024yolov10}, as TempoControl \cite{tempocontrol} does, and for \emph{single-object presence} we report GRiT \cite{wu2022grit}.\\
Because we find our three-detector majority (\cref{par:data_labeling}) more
reliable, we additionally report metrics using this aggregated detector. 

\begin{figure*}[t]
  \centering
  \includegraphics[width=\linewidth]{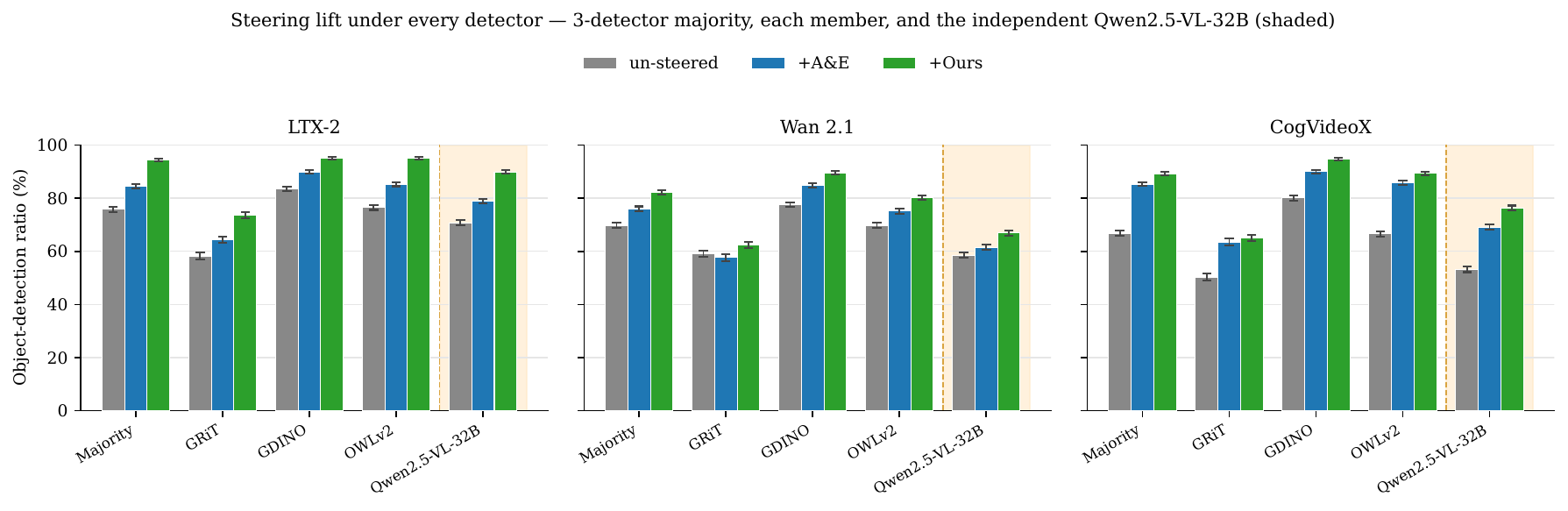}
  \caption{\textbf{Steering lift under each detector}
  ($473$ short prompts, three backbones). We re-score the same clips
  of every arm (un-steered, $+$A\&E, $+$Ours), taken from an earlier run of the comparison, under the three-detector majority,
  each member alone, and the vision--language judge
  Qwen2.5-VL-32B~\cite{bai2025qwen25vl}, used nowhere in training or labeling
  (shaded). Bars are the object-detection ratio pooled over all requested objects, so the values differ from \cref{tab:main};
  error bars are binomial SE. Under the strong independent Qwen2.5-VL, \method{}
  raises detection above both the un-steered backbone and A\&E on all three
  backbones.}
  \label{fig:detector_robustness}
\end{figure*}

\paragraph{Robustness to the detector choice.}
\label{par:detector_robustness}
Because \method{} is trained with labels from the same three-detector majority that defines our headline presence metric, one might worry that the reported gains merely reflect scoring with the detectors that supervised training.
\Cref{fig:detector_robustness} shows that they
do not: we re-score the same clips of every arm, taken from an earlier run of the comparison, across all three
backbones, under each majority member alone and under the
vision--language judge Qwen2.5-VL-32B~\cite{bai2025qwen25vl}, which was used nowhere
in our training or labeling. Under the strong
independent Qwen2.5-VL, \method{} raises detection above both the un-steered
backbone and A\&E on every backbone.

\paragraph{Label aggregation across frames.}
We use the three-detector majority for several tasks, and score a different
number of frames in each: more frames give a more accurate label at a higher
run-time cost, so we choose the number per task. For data labeling before
training and for the multi-object benchmark (\cref{tab:main}) we score $16$ frames, and for the remaining evaluations $32$; in both cases an object
counts as present when it is detected in at least $15\%$ of the scored frames.

Re-aggregating from the full per-frame detections and comparing against the full
$97$-frame labeling, both the per-object presence flip rate and the per-clip
object-count change rate fall to a few percent by the $N{=}16$/$32$ settings we
use, and reach zero only at the full $97$ frames (\cref{tab:label_flip}).

\begin{center}\footnotesize
  \captionof{table}{\textbf{Label flip rate vs.\ number of scored frames $N$.} Percentage
  of labels that differ from the full $97$-frame labeling: per-object presence
  flips and per-clip object-count ($k$) changes. The two settings we use ($16$ and $32$) are in bold.}
  \label{tab:label_flip}
  \begin{tabular}{@{}rrr@{}}
    \toprule
    $N$ & Object flip (\%) & Clip flip (\%) \\
    \midrule
    1  & 17.33 & 28.85 \\
    2  & 7.39  & 13.53 \\
    4  & 4.33  & 8.01  \\
    8  & 3.26  & 6.24  \\
    \textbf{16} & \textbf{1.99} & \textbf{3.90} \\
    \textbf{32} & \textbf{1.11} & \textbf{2.23} \\
    48 & 0.87  & 1.71  \\
    64 & 0.66  & 1.33  \\
    80 & 0.40  & 0.75  \\
    97 & 0.00  & 0.00  \\
    \bottomrule
  \end{tabular}
\end{center}

\subsection{Evaluation Metrics}

\paragraph{Temporal control.}
For object $o$, let $c_o(t)\in\{0,1\}$ specify whether the object is requested
at frame $t$, and let $d_o(t)\in\{0,1\}$ denote whether it is detected.
Comparing these sequences gives true positives (TP), false positives (FP),
true negatives (TN), and false negatives (FN). We report
\begin{equation}
\begin{aligned}
    \mathrm{Precision}
    &=
    \frac{\mathrm{TP}}{\mathrm{TP}+\mathrm{FP}},
    &
    \mathrm{Recall}
    &=
    \frac{\mathrm{TP}}{\mathrm{TP}+\mathrm{FN}}, \\
    \mathrm{Specificity}
    &=
    \frac{\mathrm{TN}}{\mathrm{TN}+\mathrm{FP}},
    &
    \mathrm{F1}
    &=
    \frac{2\,\mathrm{Precision}\,\mathrm{Recall}}
    {\mathrm{Precision}+\mathrm{Recall}} .
\end{aligned}
\label{eq:app_temporal_metrics}
\end{equation}
Recall measures whether the object is present when requested, whereas precision
penalizes objects that persist outside the requested interval. Specificity
directly measures absence outside that interval. F1 therefore prevents a method
from succeeding simply by rendering the object throughout the video.

We also evaluate temporal localization using the requested interval
$[c_s,c_e]$ and the interval $[g_s,g_e]$ spanning the first and last detections:
\begin{equation}
    \mathrm{t\text{-}IoU}
    =
    \frac{
        |[c_s,c_e]\cap[g_s,g_e]|
    }{
        |[c_s,c_e]\cup[g_s,g_e]|
    }.
    \label{eq:app_tiou}
\end{equation}
We report mean t-IoU and $\mathrm{R}@0.5$, the fraction of videos for which
$\mathrm{t\text{-}IoU}\geq0.5$. If the object is never detected, its t-IoU is $0$.
For controls requesting presence throughout the entire video, specificity is
undefined and precision is trivially $1$, so we report recall instead.

\paragraph{Video quality.}
We report the mean of seven VBench~\cite{huang2024vbench} quality dimensions: subject consistency,
background consistency, temporal flickering, motion smoothness, dynamic degree,
aesthetic quality, and imaging quality. Imaging quality (MUSIQ~\cite{ke2021musiq}) is normalized
to $[0,1]$ before averaging. \Cref{tab:vbench_dims} reports the individual
dimensions, which expose a trade-off the mean hides: on LTX-2 steering improves
subject/background consistency and imaging quality but costs $5.6$ points of
\emph{dynamic degree} against the un-steered backbone ($7.9$ against A\&E), so a
flat mean is not a flat effect. Wan~2.1 behaves differently --- its un-steered
clips are far more static (dynamic degree $28.2$ against $82.2$ on LTX-2) and
steering \emph{raises} that dimension to $50.3$, giving it the best mean of the
three Wan arms. On CogVideoX, steering improves consistency, imaging, and aesthetic quality, but lowers dynamic degree from $58.8$ to $41.0$, which places its mean $1.1$ points below the backbone.
\begin{table*}[t]
  \centering\footnotesize
  \setlength{\tabcolsep}{5pt}
  \caption{\textbf{VBench dimensions behind the single quality column} ($473$
  short prompts). The \emph{mean} column (mean7, $\pm$ SE across clips) is the
  VBench column of the main paper's presence table. \textbf{Bold} = best within a
  backbone block. On LTX-2 steering improves subject/background consistency and
  imaging quality but costs $5.6$ points of \emph{dynamic degree} against the
  un-steered backbone ($7.9$ against A\&E); on Wan~2.1 the same steering
  \emph{raises} dynamic degree ($28.2\!\to\!50.3$) from a much more static
  baseline, and takes the best mean; on CogVideoX it lowers dynamic degree ($58.8\!\to\!41.0$) and with it the mean.}
  \label{tab:vbench_dims}
  \begin{tabular}{ll ccccccc c}
    \toprule
    & & \makecell{motion\\smooth.} & \makecell{temporal\\flicker} &
        \makecell{subject\\consist.} & \makecell{background\\consist.} &
        \makecell{dynamic\\degree} & \makecell{imaging\\(MUSIQ)} &
        \makecell{aesthetic\\quality} & \makecell{VBench\\mean} \\
    \midrule
    \multirow{3}{*}{\rotatebox[origin=c]{90}{LTX-2}}
    & base         & \bf 98.8 & \bf 97.5 &     89.8 &     92.2 &     82.2 &     70.6 &     54.9 &     83.7\,{\scriptsize$\pm$0.3} \\
    & +\,A\&E      &     98.8 &     97.5 &     90.1 &     92.2 & \bf 84.5 &     70.5 & \bf 55.0 & \bf 84.1\,{\scriptsize$\pm$0.2} \\
    & +\,Ours      &     98.7 &     97.1 & \bf 92.3 & \bf 93.7 &     76.6 & \bf 71.3 &     54.8 &     83.5\,{\scriptsize$\pm$0.3} \\
    \midrule
    \multirow{3}{*}{\rotatebox[origin=c]{90}{Wan 2.1}}
    & base         & \bf 99.1 & \bf 98.7 & \bf 98.3 & \bf 97.9 &     28.2 & \bf 75.5 & \bf 61.6 &     79.9\,{\scriptsize$\pm$0.3} \\
    & +\,A\&E      &     98.8 &     98.4 &     95.6 &     96.4 &     35.2 &     71.8 &     58.3 &     79.2\,{\scriptsize$\pm$0.3} \\
    & +\,Ours      &     98.7 &     98.1 &     97.3 &     97.2 & \bf 50.3 &     74.6 &     61.3 & \bf 82.5\,{\scriptsize$\pm$0.3} \\
    \midrule
    \multirow{3}{*}{\rotatebox[origin=c]{90}{\scalebox{0.8}{CogVideoX}}}
    & base         &     98.3 & \bf 96.6 &     91.2 &     93.2 &     58.8 &     66.6 &     51.8 & \bf 79.5\,{\scriptsize$\pm$0.3} \\
    & +\,A\&E      &     98.0 &     96.1 &     89.5 &     92.1 & \bf 61.8 &     65.6 &     50.9 &     79.1\,{\scriptsize$\pm$0.3} \\
    & +\,Ours      & \bf 98.4 &     96.5 & \bf 93.5 & \bf 94.1 &     41.0 & \bf 70.6 & \bf 54.5 &     78.4\,{\scriptsize$\pm$0.3} \\
    \bottomrule
  \end{tabular}
\end{table*}

\paragraph{Statistical testing.}
All methods are evaluated on the same prompts and seeds, allowing paired
comparison per prompt. We apply an exact McNemar test to the per-prompt all-object outcome of \method{} and each baseline in \cref{tab:main}, over the prompts whose labels contain no detector tie. The gain over the un-steered backbone and over A\&E is significant on all three backbones, with $p<10^{-9}$ for every comparison except A\&E on CogVideoX ($p=0.013$).

\section{User Study}
\label{sec:suppl:userstudy}

The automatic metrics in the main paper rely on detector ensembles and
learned quality scores. To check that the improvement is visible to people, we
ran a paired-comparison study on both backbones.

\subsection{Protocol} We sample $150$ distinct prompts from the $473$-prompt
short list, stratified over the number of requested objects $k$, and render each
on both backbones, Wan~2.1 and LTX-2. For each backbone the annotator sees the
Attend-and-Excite baseline against ours: the two clips are placed side by side
--- same prompt, same seed --- in a single merged video, so the two arms cannot
drift out of sync in the annotator's browser. Every pair is pre-rendered in both
left/right orders and one is served at random, and model names never appear in
the interface. Annotators are assigned balanced blocks of $20$ clips and never
receive the same prompt on both backbones. For each pair they answer three
forced-choice questions: (1) which video shows more of the requested objects,
(2) which looks more realistic, and (3) which has smoother motion --- each with
an explicit \emph{about the same} option. \Cref{fig:userstudy_ui} shows the
interface.

We recruited $37$ volunteer annotators from two populations: computer-science
master's and PhD students, and people with no background in the field. They were
blind to which side showed which method, and the sides were randomized on every
clip.

Together they completed $622$ clip judgements and $1{,}866$ votes, a mean of
$16.8$ clips each (blocks are $20$ clips; a few sessions were abandoned
part-way). Every one of the $150$ prompts was judged on both backbones, with a mean of $2.1$ judgements per clip pair, each answering the three questions.

\begin{center}
  \includegraphics[width=0.55\linewidth]{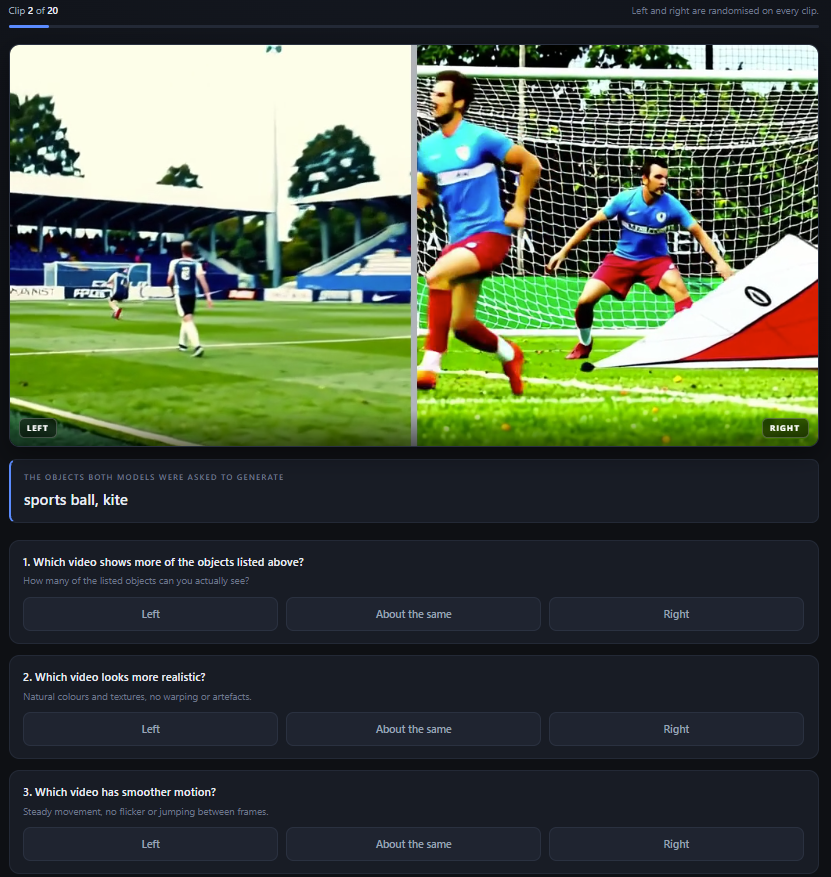}
  \captionof{figure}{\textbf{The annotation interface.} One step of the study. The two
  arms are pre-merged into a single side-by-side clip --- same prompt, same
  seed --- with the sides assigned at random and the model names never shown.
  The requested objects are listed above the three forced-choice questions,
  each of which offers an explicit \emph{about the same} option. In the pair
  shown, both clips render the sport ball, but only the right-hand
  one also renders the kite; three of the four annotators
  who saw this pair chose it as showing more of the requested objects and the
  fourth called them equal.}
  \label{fig:userstudy_ui}
\end{center}
\section{Results and Ablation Study}
\label{sec:suppl:ablation}

\noindent\includegraphics[width=\linewidth]{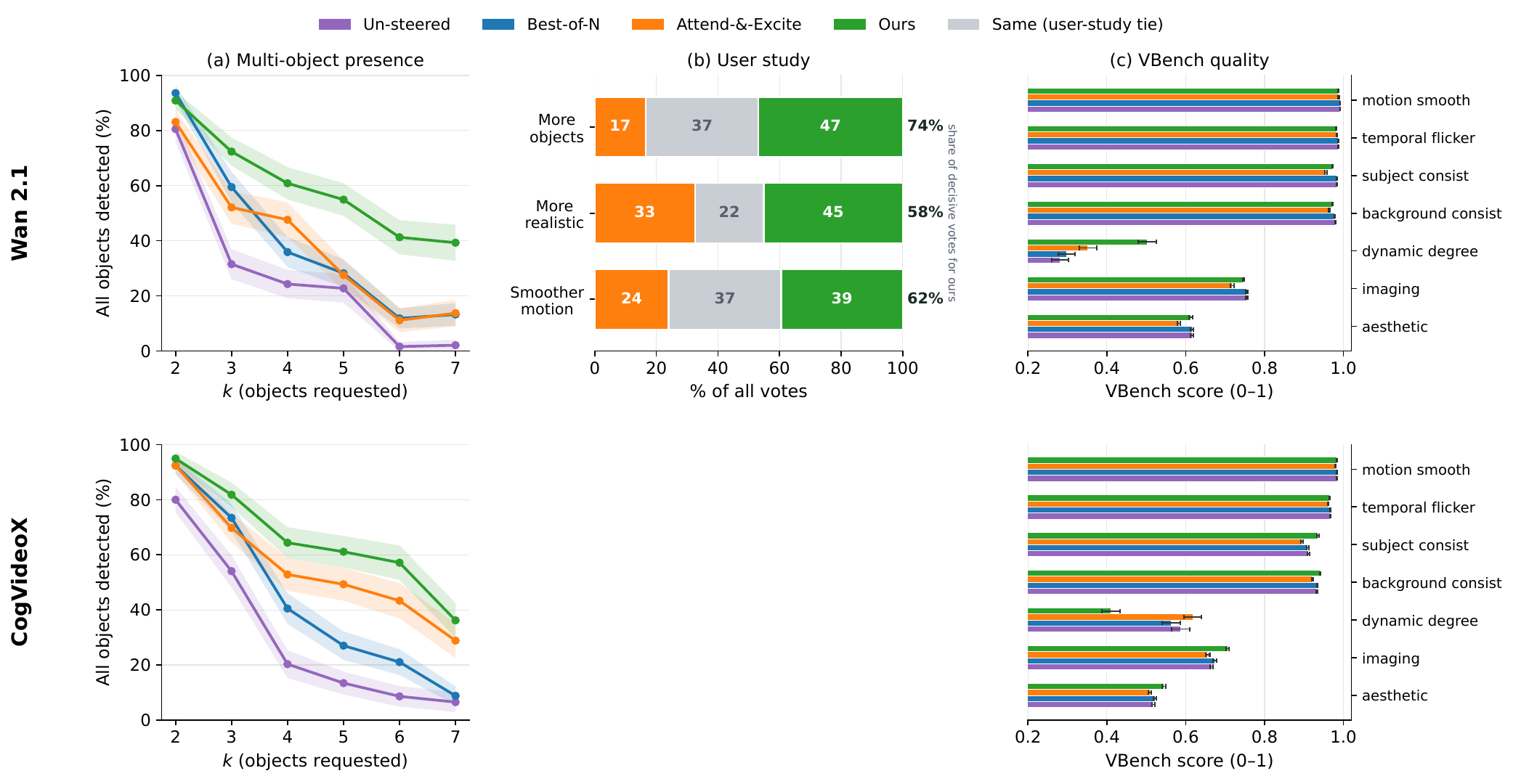}
\captionof{figure}{\textbf{Wan~2.1 and CogVideoX: presence, human preference, and quality}
(counterpart to \cref{fig:presence_by_k}, which shows LTX-2). Rows are backbones;
columns are \textbf{(a)} all-object accuracy vs.\ the number of requested objects
$k$, \textbf{(b)} human preference against Attend-and-Excite (vote composition with
the decisive-win share); \textbf{(c)} VBench quality per dimension (imaging quality as
MUSIQ$/100$).}
\label{fig:unified_appendix}

\paragraph{Per-object detection ratio.}
\Cref{tab:detratio} breaks the detection ratio down by $k$ for all four methods
on each backbone, the per-object companion to the all-objects-present split in
\cref{tab:main}: Ours leads on the average detection ratio for every backbone and
at nearly every $k$.

\noindent
\begin{minipage}[t]{0.52\linewidth}
  \centering\footnotesize
  \setlength{\tabcolsep}{5pt}
  \captionof{table}{\textbf{Detection ratio (\%) split by $k$} (473 short prompts),
  the per-object companion to the all-objects-present numbers in
  \cref{tab:main}. \textbf{Bold} = best within a backbone block.}
  \label{tab:detratio}
  \resizebox{\linewidth}{!}{%
  \begin{tabular}{ll cccccc c}
    \toprule
    & & $k$=2 & 3 & 4 & 5 & 6 & 7 & avg \\
    \midrule
    \multirow{4}{*}{\rotatebox[origin=c]{90}{Wan 2.1}}
    & base         &     90.3 &     70.8 &     72.9 &     74.8 &     65.3 &     66.1 &     73.4 \\
    & +\,Best-of-N & \bf 96.8 &     84.4 &     80.4 &     78.7 &     75.4 &     74.4 &     81.7 \\
    & +\,A\&E      &     90.9 &     80.3 &     80.6 &     76.8 &     74.7 &     75.4 &     79.8 \\
    & +\,Ours      &     95.5 & \bf 89.9 & \bf 87.7 & \bf 88.7 & \bf 85.4 & \bf 88.5 & \bf 89.3 \\
    \midrule
    \multirow{4}{*}{\rotatebox[origin=c]{90}{LTX-2}}
    & base         &     87.8 &     83.1 &     74.3 &     77.9 &     78.1 &     71.9 &     78.9 \\
    & +\,Best-of-N &     94.3 &     87.9 &     83.5 &     86.1 &     85.2 &     78.4 &     85.9 \\
    & +\,A\&E      &     87.3 &     78.9 &     75.0 &     77.5 &     78.3 &     72.8 &     78.3 \\
    & +\,Ours      & \bf 97.5 & \bf 95.6 & \bf 95.5 & \bf 93.0 & \bf 92.9 & \bf 89.0 & \bf 93.9 \\
    \midrule
    \multirow{4}{*}{\rotatebox[origin=c]{90}{CogVideoX}}
    & base         &     90.0 &     82.4 &     70.7 &     68.4 &     66.1 &     60.9 &     73.1 \\
    & +\,Best-of-N &     96.2 &     90.7 &     81.6 &     78.6 &     76.8 &     69.1 &     82.2 \\
    & +\,A\&E      &     95.6 &     89.9 &     86.4 &     87.0 &     85.3 &     79.7 &     87.3 \\
    & +\,Ours      & \bf 97.5 & \bf 93.5 & \bf 89.7 & \bf 91.4 & \bf 89.9 & \bf 85.0 & \bf 91.2 \\
    \bottomrule
  \end{tabular}}
\end{minipage}\hfill
\begin{minipage}[t]{0.45\linewidth}
  \centering
  \captionof{table}{\textbf{Conflict-aware gradient aggregation (CAGA) ablation.}
  Effect of removing CAGA---our multi-object conflict-aware gradient aggregation
  (per-object mutual agreement and unit-norm)---from \method{} on three backbones.
  We report det-all averaged over all $k{=}2$--$7$ and over the $k{=}2$--$4$ and
  harder $k{=}5$--$7$ subsets, together with VBench mean7 visual quality.
  Best per column and backbone in \textbf{bold}.}
  \label{tab:caga_ablation}
  \resizebox{\linewidth}{!}{%
  \begin{tabular}{llcccc}
    \toprule
     & & \multicolumn{3}{c}{All-object accuracy (\%)\,$\uparrow$} & VBench \\
    \cmidrule(lr){3-5}
    Backbone & Method & avg$_{2\text{-}7}$ & avg$_{2\text{-}4}$ & avg$_{5\text{-}7}$ & mean7\,$\uparrow$ \\
    \midrule
    Wan~2.1   & Ours          & $\mathbf{59.9}${\scriptsize$\,\pm2.3$} & $\mathbf{74.7}${\scriptsize$\,\pm2.8$} & $\mathbf{45.2}${\scriptsize$\,\pm3.6$} & $\mathbf{82.5}${\scriptsize$\,\pm0.3$} \\
              & Ours $-$ CAGA & $55.7${\scriptsize$\,\pm2.3$} & $74.0${\scriptsize$\,\pm2.9$} & $37.4${\scriptsize$\,\pm3.6$} & $81.8${\scriptsize$\,\pm0.3$} \\
    \midrule
    LTX-2     & Ours          & $\mathbf{75.5}${\scriptsize$\,\pm2.0$} & $\mathbf{88.7}${\scriptsize$\,\pm2.1$} & $\mathbf{62.3}${\scriptsize$\,\pm3.3$} & $83.5${\scriptsize$\,\pm0.3$} \\
              & Ours $-$ CAGA & $69.3${\scriptsize$\,\pm2.0$} & $87.2${\scriptsize$\,\pm2.2$} & $51.5${\scriptsize$\,\pm3.4$} & $\mathbf{83.9}${\scriptsize$\,\pm0.3$} \\
    \midrule
    CogVideoX & Ours          & $\mathbf{65.9}${\scriptsize$\,\pm2.2$} & $\mathbf{80.4}${\scriptsize$\,\pm2.5$} & $\mathbf{51.5}${\scriptsize$\,\pm3.5$} & $78.4${\scriptsize$\,\pm0.3$} \\
              & Ours $-$ CAGA & $55.0${\scriptsize$\,\pm2.1$} & $76.6${\scriptsize$\,\pm2.6$} & $33.4${\scriptsize$\,\pm3.3$} & $\mathbf{79.9}${\scriptsize$\,\pm0.3$} \\
    \bottomrule
  \end{tabular}}
\end{minipage}

\paragraph{Single-object presence.}
\Cref{tab:singleobj} reports the VBench \texttt{object\_class} benchmark of \cref{sec:exp} under GRiT and the three-detector majority, together with VBench quality.

\begin{table}[ht]
  \centering
  \footnotesize
  \setlength{\tabcolsep}{4pt}
  \caption{\textbf{Single-object presence.}
  VBench \texttt{object\_class} (single-object) presence, evaluated with GRiT and
  our 3-detector majority.}
  \label{tab:singleobj}
  \begin{tabular}{@{}lccc@{}}
    \toprule
    & \multicolumn{2}{c}{Presence (\%) \up}
    & VBench (\%) \up \\
    \cmidrule(lr){2-3}
    Model & 3-det. & GRiT & \\
    \midrule
    Wan~2.2
        & 88.3 & 78.5 & 79.8 \\
    LTX-2.5
        & 63.6 & 54.4 & \allbest{87.0} \\
    MiniMax-H3
        & 78.5 & 74.7 & 83.3 \\
    \cmidrule(l){1-4}
    CogVideoX-5B
        & 92.3 & 81.0 & \blockbest{78.3} \\
    \quad + A\&E
        & 90.9 & 72.2 & 75.4 \\
    \quad + Ours
        & \blockbest{\allbest{96.2}}
        & \blockbest{82.3}
        & 75.9 \\
    \cmidrule(l){1-4}
    Wan~2.1
        & \blockbest{92.2} & \blockbest{81.0} & 77.0 \\
    \quad + A\&E
        & 87.2 & \blockbest{81.0} & 76.0 \\
    \quad + Ours
        & 88.3
        & \blockbest{81.0}
        & \blockbest{79.7} \\
    \cmidrule(l){1-4}
    LTX-2
        & \blockbest{\allbest{96.2}} & 88.6 & \blockbest{82.9} \\
    \quad + A\&E
        & 91.1 & 78.5 & 82.1 \\
    \quad + Ours
        & \blockbest{\allbest{96.2}}
        & \blockbest{\allbest{89.9}}
        & 81.9 \\
    \bottomrule
  \end{tabular}
\end{table}

\begin{figure}[ht]
  \centering
  \includegraphics[width=0.906\linewidth,trim={22 308.6 0 0},clip]{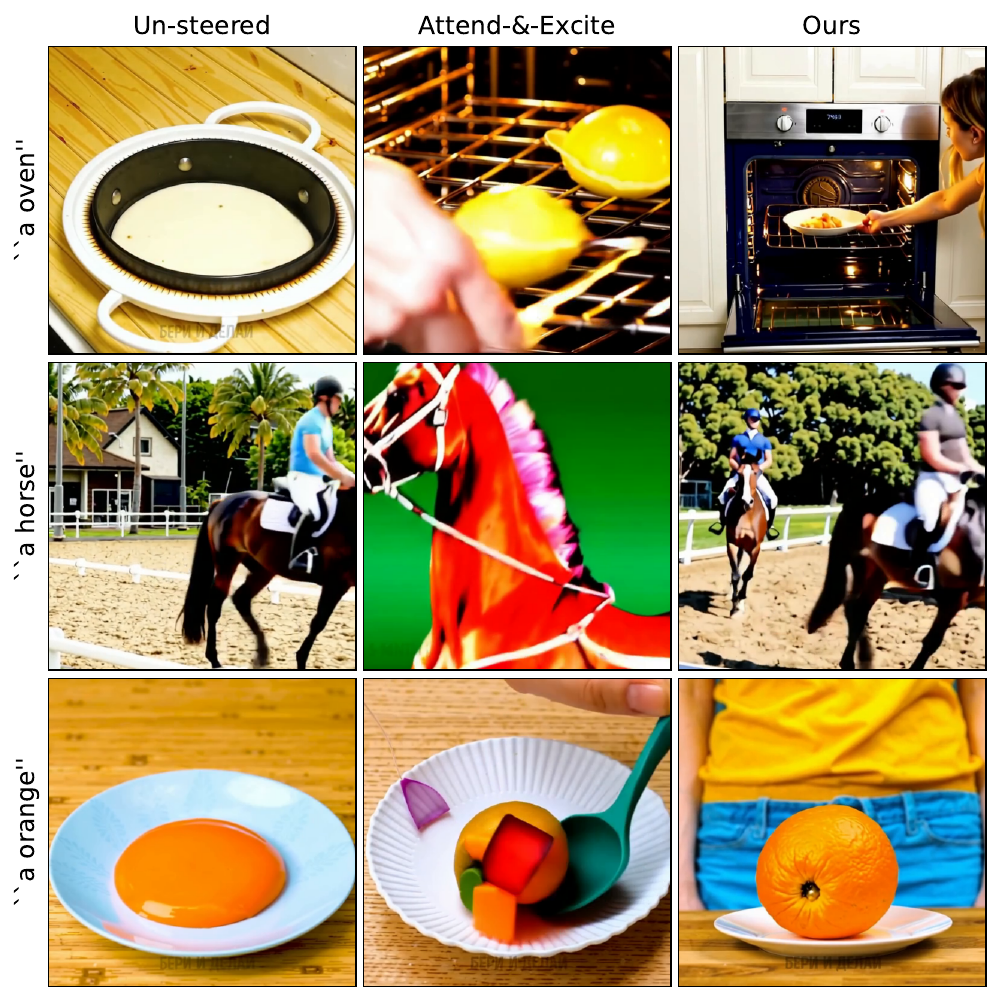}
\caption{\textbf{Single-object steering.}
Representative LTX-2 mid-frames for ``an oven.'' The un-steered backbone misses
the requested object, which \method{} renders.}
  \label{fig:singleobj}
\end{figure}

\paragraph{Conflict-aware gradient aggregation.}
\Cref{tab:caga_ablation} ablates CAGA, our multi-object conflict-aware gradient
aggregation (per-object mutual agreement and unit-norm): removing it lowers
all-object presence on every backbone, with the largest drops at high object
counts~$k$, at essentially no cost in visual quality.

\paragraph{Layer aggregation and layer count.}
\Cref{fig:layers} separates the two design choices behind the \method{} input:
how the transformer layers are combined, and how many of them are used. Over the
same features and the same validation set, \emph{learning} the per-(layer,\,head)
aggregation rather than averaging over it raises mean validation AUROC on an
$11$-layer subset from $0.811$ to $0.854$, and extending the learned aggregation
to all $48$ layers raises it further to $0.882$. Learned aggregation beats the mean at every
denoising step, and all $48$ layers beat the $11$-layer subset at every step except step $1$ ($0.801$ vs.\ $0.805$). The gaps are smallest at steps $0$ and $1$, where the latent is still
almost pure noise, and widest from step $4$ on.

\noindent
\begin{minipage}[b]{0.55\linewidth}
  \centering
  \includegraphics[width=\linewidth]{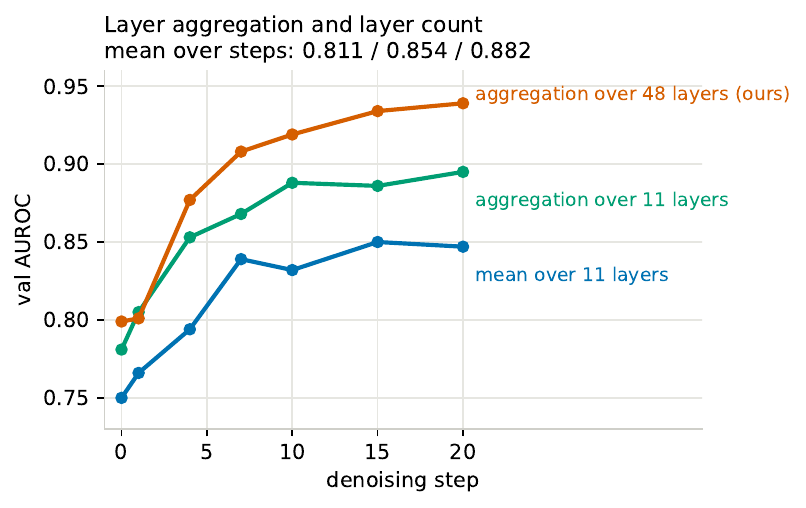}
  \captionof{figure}{\textbf{Layer aggregation and layer count.} Per-step
  validation AUROC for a uniform mean over an $11$-layer subset, our learned
  aggregation over the same $11$ layers, and our learned aggregation over all
  $48$ layers. Steps $0$--$20$, the range covered by all three runs.}
  \label{fig:layers}
\end{minipage}\hfill
\begin{minipage}[b]{0.42\linewidth}
  \centering
  \includegraphics[width=\linewidth]{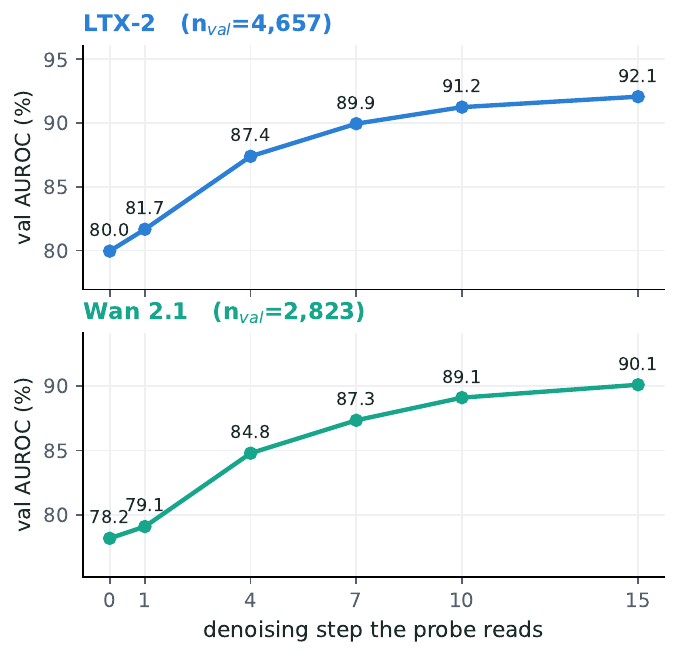}
  \captionof{figure}{\textbf{\method{} validation AUROC per denoising step.} One
  panel per backbone, one predictor per step; $n_{\text{val}}$ is that backbone's
  validation set. Both curves are shown over steps $0$--$15$, the range
  Wan~2.1 was trained on.}
  \label{fig:probe_auroc}
\end{minipage}
\section{Use of Large Language Models}
\label{sec:suppl:llm}

We disclose that large language models (LLMs) were used as general-purpose
assistive tools in the preparation of this work. Their role was limited to two
kinds of support. First, editorial assistance: proofreading, tightening and
reformatting text that the authors had already drafted, and suggesting
alternative phrasings, which the authors then accepted, rewrote or rejected.
Second, engineering assistance: writing boilerplate code for experiment
orchestration, job scheduling on our compute cluster, plotting, and the
assembly of result tables from the logs produced by our evaluation scripts.

LLMs were not used to originate the research question, design the method, or
interpret the experimental findings, and no text, number, figure or claim in
this paper was taken from a model without author verification against the
underlying experimental artifacts. The authors take full responsibility for
the entire content of the manuscript, including any part that an LLM assisted
in producing.

\section{Additional Qualitative Results}
\label{sec:suppl:qual}
\begin{figure*}[h!]
  \centering
  \includegraphics[width=\linewidth]{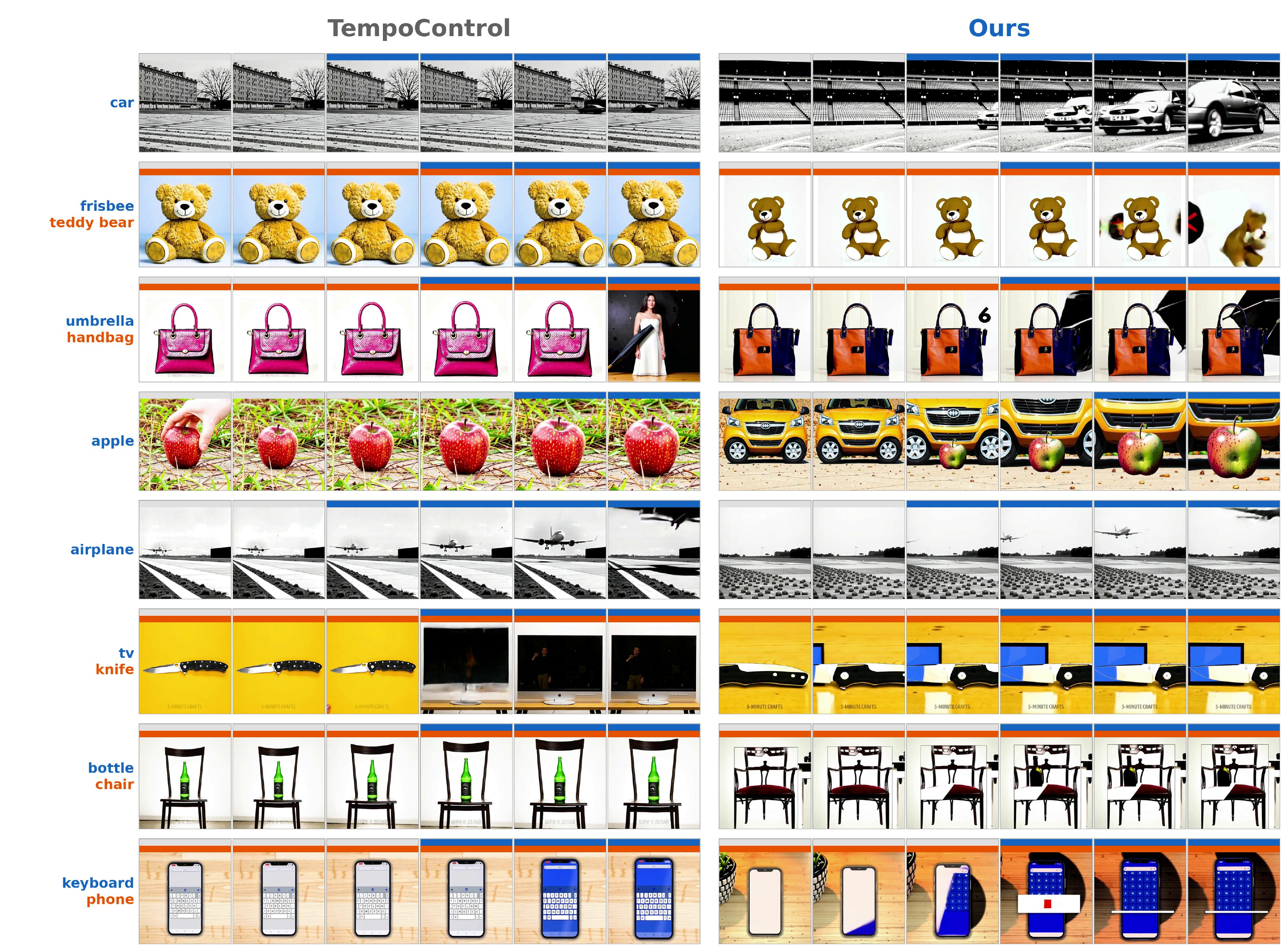}
  \caption{\textbf{Qualitative temporal steering: \method{} vs.\ TempoControl.}
  Eight LTX-2 prompts.
  Each coloured bar above a frame indicates whether the object was requested
  present (colour) or absent (light grey) at that latent timestep.
  Both columns share the same control signal; only which method succeeded
  differs.}
  \label{fig:temporal_beat_tempo}
\end{figure*}

\Cref{fig:temporal_beat_tempo} compares TempoControl and \method{} on eight LTX-2 prompts.

\end{document}